\documentclass[11pt,letterpaper]{article}

\usepackage[noend]{algorithmic}

\usepackage[margin=1in]{geometry}
\usepackage{lmodern}
\usepackage[T1]{fontenc}
\usepackage[utf8]{inputenc}
\usepackage[tbtags]{amsmath}
\allowdisplaybreaks
\usepackage{bbm}
\usepackage{amssymb,amsthm}
\usepackage{setspace} \usepackage[colorlinks,citecolor=blue,bookmarks=true,linktocpage]{hyperref}
\usepackage{url}
\usepackage[svgnames]{xcolor}
\definecolor{cblue}{rgb}{0,0,0}
\definecolor{cmagenta}{rgb}{0,0,0}
\definecolor{cmaroon}{rgb}{0,0,0}

\usepackage[style=alphabetic, maxbibnames=15, maxcitenames=15, natbib=true,
  maxalphanames=10, backend=biber, sorting=nty, backref=true]{biblatex}
\DefineBibliographyStrings{english}{backrefpage = {page},backrefpages = {pages},}

\usepackage{doi}
\usepackage{tikz} 
\usepackage{pgf,pgfplots}
\pgfplotsset{compat=1.14}
\usepgfplotslibrary{fillbetween}
\usepackage{hyphenat}
\usepackage[font=small]{caption}
\usepackage{subcaption}
\usepackage{appendix}\usepackage{thm-restate}
\usepackage{comment}

\usepackage[ruled,noend,noline]{algorithm2e}

\SetCommentSty{mycommfont}

\usepackage[capitalise,sort]{cleveref}

\usepackage{dsfont}
\usepackage{microtype,bm}
\usepackage{booktabs}
\usepackage{colortbl}
\usepackage{xcolor}
\usepackage{nicefrac}

\newtheorem{custtheorem}{Theorem}
\newtheorem*{custtheorem*}{Theorem}
\newtheorem{custlemma}[custtheorem]{Lemma}
\crefname{custlemma}{Lemma}{Lemmas}

\newtheorem{custremark}[custtheorem]{Remark}

\newtheorem{custdefinition}[custtheorem]{Definition}
\crefname{custdefinition}{Definition}{Definitions}

\newtheorem{custassumption}{Assumption} \crefname{custassumption}{Assumption}{Assumptions}
\newtheorem{custobservation}[custtheorem]{Observation}

\newcommand{\R}{\mathbb{R}}

\renewcommand{\O}{\mathcal{O}}
\newcommand{\Om}{\Omega}
\newcommand{\tO}{\widetilde{\O}}
\newcommand{\tOm}{\widetilde{\Om}}
\newcommand{\E}{\mathcal{E}}

\newcommand{\poly}{\mathrm{poly}}

\newenvironment{proofsketch}{
\begin{proof}[Proof sketch]
}
{
\end{proof}
}
\newenvironment{custproof}[1][]{
\ifthenelse{\equal{#1}{}}{
\begin{proof}
}{
\begin{proof}[Proof #1]
}
}
{
\end{proof}
}

\newcommand{\EXP}[2][]{
    \ifthenelse{\equal{#1}{}}
    {\mathbb{E}\left[#2\right]}
    {\mathop{\mathbb{E}}_{#1}\left[#2\right]}
}
\newcommand{\PRO}[2][]{
    \ifthenelse{\equal{#1}{}}
    {\mathrm{Pr}\left[#2\right]}
    {\mathop{\mathrm{Pr}}_{#1}\left[#2\right]}
}
\newcommand{\parens}[1]{\left(#1\right)}
\newcommand{\braces}[1]{\left\{#1\right\}}

\newcommand{\logp}[1]{\log\left( #1 \right)}

\newcommand{\const}{\mathrm{const}}
\newcommand{\cCommon}{c_0}
\newcommand{\tcCommon}{\widetilde{c}_0}
\newcommand{\constInvariance}{c_1}
\newcommand{\tconstInvariance}{\widetilde{c}_1}
\newcommand{\constRecImp}{c_2}
\newcommand{\constThmMain}{C}

\newcommand{\bias}{\mathrm{bias}_t}

\newcommand{\F}{\mathcal{F}}
\newcommand{\Econd}[2]{\EXP{#2 \mid \F_{#1}}}
\newcommand{\Et}[1]{\Econd{t-1}{#1}}

\newcommand{\1}[1]{\mathbbm{1}{\left\{ #1 \right\}}}

\newcommand{\abs}[1]{\left| #1 \right|}
\newcommand{\norm}[1]{\left\lVert#1\right\rVert}

\newcommand{\normSq}[1]{\norm{#1}^2}

\newcommand{\inormSq}[1]{\parens{#1}^2}
\newcommand{\innerProd}[2]{\left\langle #1, #2 \right\rangle}
\newcommand{\ceil}[1]{\left\lceil #1 \right\rceil}

\newcounter{initialIterateT}
\newcounter{initialIterateTMinusOne}
\newcommand{\initialIterateTime}{\the\value{initialIterateT}}
\newcommand{\initialIterateTimeMinusOne}{\the\value{initialIterateTMinusOne}}
\newcommand{\w}{\mathbf{w}}
\newcommand{\wzero}{\w_{\initialIterateTime}}
\newcommand{\Dom}{\R^d}
\newcommand{\fat}[1]{F(#1)}
\newcommand{\f}[1]{\fat{\w_{#1}}}
\newcommand{\fstar}{F^*}
\newcommand{\fzero}{\f{1}}
\newcommand{\ft}{\f{t}}
\newcommand{\ftplus}{\f{t+1}}

\newcommand{\gradat}[1]{\nabla F(#1)}
\newcommand{\grad}[1]{\gradat{\w_{#1}}}
\newcommand{\gradzero}{\grad{\initialIterateTime}}
\newcommand{\gradt}{\grad{t}}

\newcommand{\gradnorm}[1]{\norm{\grad{#1}}}
\newcommand{\gradtnorm}{\gradnorm{t}}
\newcommand{\gradzeronorm}{\gradnorm{\initialIterateTime}}
\newcommand{\gradsq}[1]{\normSq{\grad{#1}}}
\newcommand{\gradtsq}{\gradsq{t}}
\newcommand{\gradzerosq}{\gradsq{\initialIterateTime}}

\newcommand{\sgrad}[1]{\boldsymbol{g}_{#1}}
\newcommand{\sgradt}{\sgrad{t}}
\newcommand{\sgradzero}{\sgrad{\initialIterateTime}}
\newcommand{\sgradnorm}[1]{\norm{\sgrad{#1}}}
\newcommand{\sgradtnorm}{\sgradnorm{t}}
\newcommand{\sgradsq}[1]{\normSq{\sgrad{#1}}}
\newcommand{\sgradtsq}{\sgradsq{t}}
\newcommand{\isgrad}[1]{\boldsymbol{g}_{#1,i}}

\newcommand{\isgradsq}[1]{\inormSq{\isgrad{#1}}}
\newcommand{\isgradtsq}{\isgradsq{t}}

\newcommand{\sigmaZero}{\sigma_0}
\newcommand{\sigmaOne}{\sigma_1}

\newcommand{\etaat}[1]{\eta_{#1}}
\newcommand{\etat}{\etaat{t}}
\newcommand{\tetaat}[1]{\tilde{\eta}_{#1}}
\newcommand{\tetat}{\tetaat{t}}
\newcommand{\tetalb}{\underline{\tetaat{T}}}

\newcommand{\itetaat}[1]{\tilde{\eta}_{#1,i}}
\newcommand{\itetat}{\itetaat{t}}

\newcommand{\Sg}{S_{\mathrm{good}}}
\newcommand{\Scomp}[1]{S_{[#1]}^{\mathrm{comp}}}
\newcommand{\Sel}[1]{S_{[#1]}^{\mathrm{eligible}}}
\newcommand{\Sc}{S^{\mathrm{comp}}}

\newcommand{\tSg}{\widetilde{S}}
\newcommand{\tSgc}{\tSg^c}
\newcommand{\tb}[1]{\tau_{[#1]}^{\mathrm{bad}}}

\newcommand{\ncomp}{n_{\mathrm{comp}}}

\newcommand{\stepBnd}{\beta_{1}}
\newcommand{\decayBnd}{\beta_{2}}
\newcommand{\stepBndAGNorm}{\eta}
\newcommand{\decayBndAGNorm}{\eta^2}

\newcommand*\samethanks[1][\value{footnote}]{\footnotemark[#1]}

\author{Matthew Faw\thanks{Equal contribution} \thanks{Department of Electrical and Computer Engineering, University of Texas at Austin.\newline{\tt\{\{\href{mailto:matthewfaw@utexas.edu}{matthewfaw},\href{mailto:isidoros_13@utexas.edu}{isidoros\_13},\href{mailto:constantine@utexas.edu}{constantine},\href{mailto:sanjay.shakkottai@utexas.edu}{sanjay.shakkottai}\}@utexas.edu, \href{mailto:mokhtari@austin.utexas.edu}{mokhtari}@austin.utexas.edu\}}}
\and
Isidoros Tziotis\samethanks[1] \samethanks[2]
\and
Constantine Caramanis\samethanks[2]
\and
Aryan Mokhtari\samethanks[2]
\and
Sanjay Shakkottai\samethanks[2]
\and
Rachel Ward\thanks{Department of Mathematics, University of Texas at Austin. {\tt \{\href{mailto:rward@math.utexas.edu}{rward}@math.utexas.edu}\}}
}

\title{The Power of Adaptivity in SGD: Self-Tuning Step Sizes with Unbounded Gradients and Affine Variance}

\date{}

\makeatletter
\let\original@algocf@latexcaption\algocf@latexcaption
\long\def\algocf@latexcaption#1[#2]{\@ifundefined{NR@gettitle}{\def\@currentlabelname{#2}}{\NR@gettitle{#2}}\original@algocf@latexcaption{#1}[{#2}]}
\makeatother

\begin{document}
\maketitle

\begin{abstract}We study convergence rates of AdaGrad-Norm as an exemplar of adaptive stochastic gradient methods (SGD), where the step sizes  change based on observed stochastic gradients, for minimizing non-convex, smooth objectives. Despite their popularity, the analysis of adaptive SGD lags behind that of non adaptive methods in this setting. Specifically, all prior works rely on some subset of the following assumptions: (i) uniformly-bounded gradient norms, (ii) uniformly-bounded stochastic gradient variance (or even noise support), (iii) conditional independence between the step size and stochastic gradient. In this work, we show that AdaGrad-Norm exhibits an order optimal convergence rate of $\O(\nicefrac{\poly\log(T)}{\sqrt{T}})$ after $T$ iterations under the same assumptions as optimally-tuned non adaptive SGD (unbounded gradient norms and affine noise variance scaling), and crucially, without needing any tuning parameters. We thus establish that adaptive gradient methods exhibit order-optimal convergence in much broader regimes than previously understood.

\end{abstract}

\newpage
\tableofcontents
\newpage

\section{Introduction}
\label{sec:intro}

Due to its simplicity, an enormous amount of literature, starting by \citet{RM51}, has sought to understand convergence guarantees for variants of stochastic gradient descent (SGD):
\begin{align*}
    \w_{t+1} = \w_t - \etat \sgradt,
\end{align*}
for minimizing a function $\fat{\cdot}$ using stochastic gradients $\sgradt$ and a step size schedule $\etat$. When the (non-convex) objective function is smooth (i.e., has $L$-Lipschitz-continuous gradients) and the stochastic gradients are unbiased and have affine variance\footnote{While the proof of convergence under affine variance is not given explicitly in \citep{GL13}, by slightly modifying the step size choice, the analysis given in this work continues to hold with no additional modifications. Indeed, this observation is made explicitly by \citet[Theorem 4.8]{BCN18}.}, i.e.,
\begin{align}\label{eq:introAssump}
    \EXP{\sgrad{}} = \grad{}
    \quad \text{and} \quad
    \EXP{\normSq{\sgrad{} - \grad{}}} \leq \sigmaZero^2 + \sigmaOne^2\gradsq{},
\end{align}
then it is well-known that SGD with a properly-tuned step size (depending on $L$ and $\sigmaOne$) converges to a first-order stationary point with error $\O(\nicefrac{1}{\sqrt{T}})$ after $T$ iterations \citep{GL13,BCN18}. Moreover, \cite{ACDFSW19} showed this rate is tight under these assumptions. 

Given these results, it is natural to ask if knowledge of $L$ and $\sigmaOne$ is {necessary} to obtain this optimal rate of convergence. Indeed, this has been the motivation for adaptive step size algorithms such as AdaGrad-Norm, where for any parameters $\eta,b_0 > 0$, the step size, $\etat$, is given by
\begin{equation}\tag{AG-Norm}\label{eq:alg}
    \etat = \frac{\eta}{b_t}, \quad \text{where} \quad b_t^2 = b_0^2 + \sum_{s=1}^t \sgradsq{s} = b_{t-1}^2 + \sgradtsq.
\end{equation}

\citet{WWB19} showed that AdaGrad-Norm
enjoys a $\O(\nicefrac{\log(T)}{\sqrt{T}})$ convergence rate even when neither $L$ nor $\sigma_0$ is used to tune the step size-schedule. However, their analysis only holds when $\sigmaOne = 0$
and the gradients are {uniformly} upper-bounded -- an assumption which is violated even by strongly convex functions such as $\fat{\w} = \normSq{\w}$. In fact, \citep[Section~4]{LO19} suggests that, due to the correlation between $\etat$ and $\sgradt$ in the standard AdaGrad-Norm, the assumption that the gradients are uniformly-bounded might be {necessary} to prove their convergence guarantee. Although some works on similar adaptive SGD algorithms do not require the gradients to be uniformly upper-bounded \citep{LO19,LO20}, their analysis only holds when the step-size $\etat$ is (conditionally) independent of the current stochastic gradient $\sgradt$, and require subgaussian noise (a condition which forces $\sigmaOne = 0$). However, disentangling $\etat$ from $\sgradt$ is detrimental to the normalization scheme, rendering these methods crucially {dependent} on the knowledge of the Lipschitz constant $L$ for determining their step size.

Extending these results from the bounded variance setting ($\sigmaOne = 0$) to the affine variance setting is important. Indeed, results that hold only for the case of bounded variance effectively require that one has noiseless access to gradients when their magnitudes are large (see Remark~\ref{rem:affinevar} for more discussion). As opposed to the non-adaptive SGD setting where this extension is immediate (discussed above), in AdaGrad-Norm (and more generally, in adaptive methods), the bias introduced by the correlation between $\etat$ and $\sgradt$ causes this additional variance to be significantly more problematic.

\subsection{Contributions, Key Challenges and the Main Insights}
\label{sec:overview}

We show that AdaGrad-Norm
converges to a first-order stationary point with error $\O(\nicefrac{\poly\log(T)}{\sqrt{T}})$ after $T$ iterations under the same noise assumptions as well-tuned SGD (stochastic gradients are unbiased, with affine variance, as in \eqref{eq:introAssump}). Thus, we achieve a convergence rate with optimal dependence on $T$ up to polylogarithmic factors \citep{ACDFSW19},
even when the step-size sequence is chosen without knowledge of $L,\sigmaZero$, or $\sigmaOne$. In a sense, this establishes a ``best of both worlds'' result for adaptive SGD methods, showing that they can converge at the same rate (up to logarithmic factors) as in \citep{GL13} without {any} hyperparameter tuning of the step-size sequence.
Our results show that neither the assumption of uniformly-bounded gradients nor the assumption of uniformly-bounded variance is necessary; thus, adaptive gradient methods exhibit robust performance in much broader regimes than what has been established by prior studies.

Our analysis must overcome two main challenges: (i) possibly {unbounded} gradients, and (ii) an additional bias term introduced by {affine} variance. Prior work avoided or circumvented these challenges via additional assumptions. Our work requires several new insights that we believe may be of independent interest. Furthermore, as we state in Remark~\ref{ada_grad_remark}, these insights are broadly applicable to related adaptive algorithms such as coordinate-wise AdaGrad. We outline these below.

\paragraph{Main Challenge 1: Unbounded gradients.} Prior work by \citet{WWB19}, under uniformly bounded gradients and  uniformly bounded variance assumptions, introduce a proxy $\tetat$ for the step size in \eqref{eq:alg}. Unlike $\eta_t$ (the true step size), this proxy is decorrelated from $\sgradt$. Furthermore, this proxy scales inversely to (the square root of) the sum of gradients. The boundedness assumption is used to deterministically bound each individual gradient term in the sum, and thus derives a lower-bound of $\EXP{{\tetat}}= \Om(\nicefrac{1}{\sqrt{T}})$. 
This directly leads to a convergence rate of $\tO(\nicefrac{1}{\sqrt{T}})$ to a first-order stationary point in their context. Without the bounded gradient and thus, bounded variance assumptions, however, it is unclear if $\EXP{{\tetat}}$ scales as $\widetilde{\Omega}(\nicefrac{1}{\sqrt{T}})$. Instead of assuming a uniform, deterministic bound on each summand as in the prior approach, we develop techniques of independent interest that permit us to directly bound this sum in expectation.

{\textit{Key Insight 1: Recursively-improving inequalities.}} 
We identify two properties satisfied by AdaGrad-Norm (as well as related adaptive algorithms) -- bounded iterate steps and norm-squared step decay -- which allow us to derive an initial lower bound of $\tetat = \Om(\nicefrac{1}{\poly(T)})$ which holds with sufficiently high probability, and a corresponding upper bound on the sum of the gradients of $\sum_{t\in[T]} \gradtsq = \O(T^2 \log (T))$. While this polynomial
bound is too loose to result in any convergence rate, it does provide a starting point.
Our key technical approach here is a recursion, where in each iteration, we improve both these bounds using a result that shows their product is controlled by an invariant upper bound (\cref{lem:informalMainSummedBoundAffineInterpreted}). By infinitely  recursing this argument, so that constants or logarithmic factors do not ``blow up,'' we obtain an order-optimal bound directly on the expected sum of gradients, {eliminating} the need for a uniform upper bound on individual gradients.

\paragraph{Main Challenge 2: Additional bias from affine variance.}

In the affine variance setting, 
the expected difference in function value between consecutive time steps is bounded as:
\begin{equation}\label{eq:challenge}
    \frac{\tetat}{2}\parens{1-\sigmaOne\ \bias}\gradtsq{} \!
    \leq \Et{\ft{} - \ftplus{}}
    + \const \ \Et{\etat^2 \sgradtsq},
\end{equation}
where $\const$ is a constant which scales with $\sigmaZero$ and $L$. 
Whenever $\bias > \nicefrac{1}{\sigmaOne}$, then the ``negative drift'' term from the bounded variance case, $-\tetat \cdot \gradtsq$, becomes {positive}, making the derivation of the invariant upper bound identified above (in {\em Key Insight 1}, \cref{lem:informalMainSummedBoundAffineInterpreted}) a serious challenge.
The presence of this $\bias$ is the reason that prevents the analysis from the uniformly-bounded variance case to directly extend to the affine variance framework, as happens in the standard SGD analysis of \citet{GL13} by simply scaling down the step size by $\nicefrac{1}{(1+\sigmaOne^2)}$.

\textit{Key Insight 2: Focus on the ``good'' times.}
To handle this $\bias$, we first restrict our analysis to a subset of time steps, $\Sg = \{t \in [T] : \bias \leq \nicefrac{1}{2 \sigmaOne}\}$, which we refer to as the ``good'' time steps. Intuitively, these are the time steps during which the $\bias$ term is sufficiently small. As it turns out, the overwhelming majority of time steps are, in fact, ``good,'' as shown in \cref{lem:informalGoodSet}.

\textit{Key Insight 3: Compensating for the ``bad'' times.}
Although the overwhelming majority of time steps are ``good,'' in order to get a convergence rate that depends on $\fzero - \fstar$, we still have to reason about the ``bad'' time steps in $\Sg^c$. As it turns out, if the gradient at {even one} of these bad times is large (say, $\gradtsq = T^{\Omega(1)}$) then our upper bound on $\ftplus - \ft$ is prohibitively large, presenting a serious challenge for the convergence analysis. We circumvent this issue using a novel approach that assigns nearby (in terms of time) ``good'' times to every ``bad'' one, thereby mitigating the effects of ``bad'' time steps in the analysis. This compensation insight, formalized in \cref{lem:compensationInsight}, coupled with the fact that ``most'' time-steps are typically ``good,'' allows us to overcome the {bias} term introduced by the affine variance scaling. 

 \textbf{Related Work}.
\citet{GL13} were the first to study the convergence of SGD for opimizing a non-convex, smooth objective function. They proved that a properly-tuned SGD converges to a first-order stationary point at rate $\O(\nicefrac{1}{\sqrt{T}})$, if the step sizes are chosen as $\etat = \min\braces{\nicefrac{1}{(1+\sigmaOne^2)L},\nicefrac{\widetilde{D}}{\sigmaZero \sqrt{T}}}$ for a constant $\widetilde{D}>0$. {\color{black} Further, \citet{ACDFSW19} proved that the $\O(\nicefrac{1}{\sqrt{T}})$  rate is unimprovable for any algorithm with only first-order oracle access, assuming the function is non-convex, smooth, and the stochastic gradients are unbiased with bounded variance.}

The original AdaGrad algorithm was proposed simultaneously by \citet{DHS11,MS10} whereas \citet{SM10} were the first to consider a variant of AdaGrad referred to as AdaGrad-Norm. \citet{WWB19} analyzed AdaGrad-Norm for minimizing a smooth, non-convex function with uniformly-bounded gradients. 
{\color{black} They showed that} AdaGrad-Norm converges at essentially the same rate as SGD, but without the need to know the smoothness constant (albeit under the restrictive assumption that the gradients are uniformly upper-bounded). In a simultaneous work, \citet{LO19} studied a variant of AdaGrad-Norm where step size $\etat$ is conditionally independent of the current stochastic gradient $\sgradt$, unlike in the standard AdaGrad setting. They provided a similar convergence guarantee without needing a uniform upper-bound on the stochastic gradients, but requiring that the noise have bounded support and additionally requiring knowledge of the smoothness parameter $L$ to tune their step sizes. In a followup work \citep{LO20}, the same authors proved \emph{high-probability} convergence of a class of adaptive algorithms (including their variant of AdaGrad-Norm, as well as coordinate-wise AdaGrad with momentum) under the assumption of subgaussian noise. Note that, like the earlier result, their step sizes needed to be tuned with knowledge of the smoothness parameter, and further needed to be conditionally independent of the current gradient. {\color{black}\citet{KLC22} established high probability results for AdaGrad without knowledge of the smoothness parameter in the bounded variance regime, assuming that the norm of the gradients are uniformly upper-bounded (i.e., the objective function is Lipschitz). They were further able to remove the Lipschitz assumption, but only when in addition to bounded variance, the noise of the stochastic gradients is subgaussian.}
\citet{GG20} studied the asymptotic convergence of AdaGrad (as well as and RMSProp), where their analysis requires uniform gradient bounds as well as uniform bounds on the $2$nd and $4$th moments of the gradient noise.
{\color{black}
Very recently, 
\citet{JXH22} established asymptotic almost-sure convergence of the AdaGrad-Norm iterates to first-order stationary points. Unlike our work, they do not provide rates of convergence, and their focus on asymptotics makes their analysis and results significantly different.}
\citet{ZSJSL18} studied a weighted version of coordinate-wise AdaGrad with momentum, where they assumed the gradients were uniformly bounded. \citet{DBBU20} later improved upon these results with respect to the dependence on the momentum parameter.

Several recent works have studied the convergence of other adaptive algorithms, all of which are based on the assumption of uniformly-bounded stochastic gradients.
{\color{black} For instance, \citet{KLBC19} developed an adaptive, accelerated algorithm that achieves optimal rates in the constrained, convex (smooth and non-smooth) regime, without knowledge of the smoothness or noise parameters.} \citet{CLSH18} studied the convergence of a class of Adam-like algorithms (originally introduced by \citet{KB14}).  
Later, building on the results of \citet{WWB19}, \citet{DBBU20} improved on this analysis of Adam with respect to the dependence on the momentum parameter and range of valid hyperparameters. \citet{GXYJY21} provide an alternate analysis of a class of Adam-like algorithms for different momentum parameter scaling. \citet{SMBM21} studied ``delayed'' versions of Adam (as well as a new algorithm they called AvaGrad), which makes the step sizes $\etat$ conditionally independent of the current stochastic gradient, $\sgradt$.

 \section{Preliminaries}
\label{sec:preliminaries}

We study the convergence of stochastic gradient descent with adaptively chosen step sizes for minimizing a non-convex, smooth function $\fat{\cdot}$ over {unbounded} domain $\Dom$ with $\fstar = \inf_{\w\in\R^d}\fat{\w} > -\infty$. In our context, {adaptive} step sizes are those which \emph{depend} on the current stochastic gradient, as well as, potentially, those from past iterates.
We focus on the AdaGrad-Norm algorithm \eqref{eq:alg}, although our arguments readily extend to the coordinate-wise AdaGrad case (albeit, at a cost of additional dependence on the dimension). 
We denote
$\F_{t} = \sigma\braces{\wzero, \sgradzero,\ldots,\w_{t},\sgradt,\w_{t+1}}$
as the sigma algebra generated by the observations of the algorithm after observing the first $t$ stochastic gradients, and use $\norm{\cdot}$ to denote the $\ell_2$ norm. We assume the following throughout the paper.
\begin{custassumption}[Unbiased gradients]\label{assump:unbiasedGrad}
For each time $t$, the stochastic gradient, $\sgradt$, is an unbiased estimate of $\gradt$, i.e.,
$\Et{\sgradt} = \gradt$.
\end{custassumption}
\begin{custassumption}[Affine variance]\label{assump:affineVariance}
For fixed constants $\sigmaZero,\sigmaOne \geq 0$, the variance of the stochastic gradient $\sgradt$ at any time $t$ satisfies
$\Et{\normSq{\sgradt - \gradt}} \leq \sigmaZero^2 + \sigmaOne^2 \gradtsq$.
\end{custassumption}

\begin{custremark} \label{rem:affinevar} \textbf{(Motivation for Affine Variance)} This scaling is important for  machine learning applications with feature noise (including missing features) \citep{Fuller09,khli20}, in robust linear regression \citep{xu2008robust}, and generally whenever the model parameters are multiplicatively perturbed by noise (e.g., a multilayer network, where noise from a previous layer multiplies the parameters in subsequent layers). More broadly, restricting to bounded variance (i.e., assuming $\sigmaOne^2 = 0$) is equivalent to assuming ``noiseless'' access to the gradient when the magnitude of the gradient grows (e.g., a strongly convex function); this is because the stochastic gradient is an arbitrarily small perturbation of the true gradient in this regime. Finally, as discussed earlier, the analysis for non adaptive SGD is essentially unaffected by affine variance \citep{BCN18}.
\end{custremark}

Since $\Et{\innerProd{\sgradt - \gradt}{\gradt}} = 0$, we note that \cref{assump:unbiasedGrad,assump:affineVariance} imply that
\begin{align}\label{eq:sgradBound}
    \Et{\sgradtsq} \leq \sigmaZero^2 + (1+\sigmaOne^2) \gradtsq.
\end{align}
Further, we will assume that the function $\fat{\cdot}$ is $L$-smooth:

\begin{custassumption}[$L$-smoothness]\label{assump:smooth}
    The function $\fat{\cdot}$ is \emph{$L$-smooth}, i.e., has $L$-Lipschitz continuous gradients. That is, for every $\w,\w'\in\Dom$,
$\norm{\gradat{\w} - \gradat{\w'}}
        \leq L \norm{\w - \w'}$.
\end{custassumption}

A key property of AdaGrad-Norm is that the step-size sequence is tightly controlled: \begin{align}\label{eq:stepSizeControl}
    \norm{\w_{t+1}-\w_t}\leq \eta
    \quad\text{ and }\quad
    \sum_{t\in[T]} \normSq{\w_{t+1}-\w_t}\leq\eta^2\log(\nicefrac{b_T^2}{b_0^2}).
\end{align}
In fact, variations of this observation have been noted for a number of AdaGrad variants \citep{WWB19,DBBU20}. While simple, it is crucially important to our analysis, since, taken together with \cref{assump:smooth}, it implies that the gradient at time $t$ scales at most \emph{polynomially} in $t$.

\begin{custlemma}[Polynomial control of gradients (informal statement of \cref{lem:trivialGradBound,lem:logGradBound})]\label{lem:polyGradBound}
Consider any times $t_1 \leq t_2\in [T]$ during a run of algorithm \eqref{eq:alg}. Then, deterministically,
\begin{align*}
    \abs{\gradnorm{t_2} - \gradnorm{t_1}} \leq \eta L (t_2 - t_1).
\end{align*}
Moreover, with probability at least $1-\delta$, the following bound also holds
\begin{align*}
    \abs{\gradnorm{t_2} - \gradnorm{t_1}} \leq \eta L \sqrt{(t_2 - t_1)\log(\nicefrac{\poly(t_2)}{\delta})}.
\end{align*}
\end{custlemma}
As a consequence of \cref{lem:polyGradBound}, we derive $\sum_{t\in[T]} \gradtsq = T(\gradzeronorm + \eta L T)^2 = \O(T^3)$ deterministically, and an analogous bound of $\O(T^2\log(\nicefrac{T}{\delta}))$  with probability $1-\delta$.
Of course, \cref{lem:polyGradBound} only gives a much weaker control over $\gradtsq$ than a uniform bound, and has not (to the best of our knowledge) been previously exploited. However loose, this bound nonetheless is one of the key steps to removing the uniform gradient bound, and may be of independent interest (e.g., useful for refining the convergence rates for strongly convex problems).

As mentioned earlier, a key difficulty in analyzing {adaptive} algorithms is the {bias} introduced by the correlation between the step size $\etat$ and the stochastic gradient $\sgradt$ at each time $t$. To analyze the convergence of such algorithms, it is useful to introduce the following ``decorrelated'' step size.
\begin{custdefinition}[Decorrelated step sizes]\label{def:stepSizeProxy}
The \emph{decorrelated} step size ``proxy'' at time $t$, which is independent (conditioned on the history $\F_{t-1}$) of $\sgradt$, is denoted by $\tetaat{t}$ and defined as  
 $$\tetaat{t} := \frac{\eta}{\sqrt{b_{t-1}^2 + (1+\sigmaOne^2)\gradtsq + \sigmaZero^2}}.$$ 
\end{custdefinition}
Notice that $\tetat$ is the natural lower bound on $\Et{\etat}$ by applying Jensen's inequality.

 \section{Motivating the Proof}
\label{sec:challenges}
We have discussed the two main challenges in Section \ref{sec:overview}: unbounded gradients and affine variance. Now that we have the required mathematical definitions from Section \ref{sec:preliminaries}, we discuss these challenges in more detail.
Adaptive stochastic gradient methods exhibit two difficulties not present in the non-adaptive regime:
(i) Since the step size $\etat$ depends on the trajectory of stochastic gradients, one must argue about the scaling of these stochastic gradients, and (ii) the step size is {correlated} with the current gradient, $\sgradt$, as well as the past gradients. These manifest themselves as follows: by $L$-smoothness (\cref{assump:smooth}) and the AdaGrad-Norm algorithm \eqref{eq:alg}, we have that
\begin{align}\label{eq:standardAnalysis}
    \etat \gradtsq
    \leq \ft - \ftplus - \etat\innerProd{\gradt}{\sgradt - \gradt} + \frac{L\etat^2}{2}\sgradtsq.
\end{align}
When $\etat$ and $\sgradt$ are conditionally independent, then the inner product term above is {mean-zero}. As a consequence, as long as the step size $\etat \leq \nicefrac{1}{L(1+\sigmaOne^2)}$, \eqref{eq:standardAnalysis} immediately implies that
\begin{align}\label{eq:standardAnalysisExpSummed}
    \EXP{\sum_{t\in [T]} \frac{\etat}{2} \gradtsq} \leq \fzero - \fstar + \frac{L\sigmaZero^2}{2}\sum_{t\in [T]} \etat^2.
\end{align}
Moreover, {\color{cblue}if} $\etat = \Omega(\nicefrac{1}{\sqrt{T}})$,
a $\O(\nicefrac{1}{\sqrt{T}})$ obtaining the convergence rate is immediate (see \citep{GL13,BCN18} for details).\
In contrast, in the {adaptive} setting, the inner product term of \eqref{eq:standardAnalysis} may {no longer} be mean-zero, since $\etat$ {depends} on $\sgradt$.\ While \citet{LO19} circumvented this issue by studying a step-size sequence which depends on the {past} but {not} current gradient, \citet{WWB19} and \citet{DBBU20} analyzed adaptive gradient methods by introducing (for the sake of analysis) a step-size proxy (identical to \cref{def:stepSizeProxy} for $\sigmaOne=0$), 
$   \tetat = \nicefrac{\eta}{\sqrt{b_{t-1}^2 + \gradtsq + \sigmaZero^2}},$
which {is} conditionally independent of $\sgradt.$
Using that, \eqref{eq:standardAnalysis} can be rewritten as
\begin{align}\label{eq:standardAnalysisRewritten}
    \tetat \gradtsq
    &\leq \Et{\ft - \ftplus} + \Et{(\tetat - \etat)\innerProd{\gradt}{\sgradt}}\nonumber\\
    &\quad+ \Et{\frac{L\etat^2}{2}\sgradtsq}.
\end{align}
As noted in prior work, one can show that $\EXP{\sum_{t\in[T]} \etat^2 \sgradtsq} = \O(\log(T))$  ({\color{black} note that this need not be true in the non-adaptive setting}; see \cref{cor:logBound} for a proof in our setting) and thus, for the remainder of this discussion, we focus only on the remaining terms of \eqref{eq:standardAnalysisRewritten}.

\paragraph{{Unbounded Gradients: Lower-bounding the step size.}} Although in the non-adaptive setting, we could simply choose $\etat = \Omega(\nicefrac{1}{\sqrt{T}})$, in the adaptive regime it is no longer obvious that such a condition holds. One may observe, however, that by Jensen's inequality and \cref{def:stepSizeProxy}
\begin{align}\label{eq:lowerBoundStepSize}
    \EXP{\etat} \geq \EXP{\tetat} \geq \frac{\eta}{\sqrt{b_0^2 + T \sigmaZero^2 + (1+\sigmaOne^2)\EXP{\sum_{s\in[t]} \gradsq{s}}}}.
\end{align}
As discussed in Section~\ref{sec:overview}, it should be clear by observing \eqref{eq:lowerBoundStepSize} that the reason prior studies \citep{WWB19,GG20,ZSJSL18,DBBU20}
assumed a uniform upper bound on the gradients is to bound the denominator in \eqref{eq:lowerBoundStepSize}. This  allows one to conclude that both $\etat$ and $\tetat$ scale as $\Omega(\nicefrac{1}{\sqrt{T}})$ in expectation. 
Since our setting is one where neither the gradients nor the variances are uniformly bounded, new techniques are required to get around this challenge.

\paragraph{{Affine Variance: Upper-bounding the bias.}}
The bias term in \eqref{eq:standardAnalysisRewritten} presents another difficulty in analyzing the rate of convergence in the adaptive setting. Specifically, in the affine variance setting
\begin{equation}\label{eq:adaptiveAnalysisAffineBias}
    \Et{(\tetat - \etat)\innerProd{\gradt}{\sgradt}}
    \leq \frac{\tetat}{2}\parens{1 + \sigmaOne\ \! \bias}\gradtsq+ \frac{2 \sigmaZero}{\eta} \Et{\etat^2 \sgradtsq},
\end{equation}
where $\bias := 4 \sqrt{\Et{\nicefrac{\sgradtsq}{(b_{t-1}^2+\sgradtsq)}}}$ is the additional bias introduced by the affine variance scaling (see \cref{lem:analogueAffine}).
Notice that in the bounded variance setting (i.e., $\sigmaOne=0$), \eqref{eq:adaptiveAnalysisAffineBias} corresponds precisely to the bound obtained by \citet{WWB19} which was used to derive
\begin{align}\label{eq:standardAnalysisRewrittenExpSummed}
    \EXP{\sum_{t\in [T]} \frac{\tetat}{2} \gradtsq} \leq \fzero - \fstar + 
    \cCommon \log(\poly(T)),
\end{align}
where $\cCommon = 2\sigmaZero \eta + \nicefrac{L\eta^2}{2}$.
This inequality is analogous to \eqref{eq:standardAnalysisExpSummed} and, combined with the lower bound $\EXP{{\tetat}}= \tOm(\nicefrac{1}{\sqrt{T}})$, immediately leads to the desired convergence rate.
When $\sigmaOne \leq \nicefrac{1}{8}$, \eqref{eq:adaptiveAnalysisAffineBias} takes essentially the same form as \eqref{eq:standardAnalysisRewrittenExpSummed}, since, deterministically, $\Et{\nicefrac{\sgradtsq}{(b_{t-1}^2 + \sgradtsq)}}\leq 1$.
When $\sigmaOne \geq \nicefrac{1}{4},$ however, the first term of \eqref{eq:adaptiveAnalysisAffineBias} can potentially be quite large\footnote{While one could control this term using a batch size of $\Omega(\sigmaOne^2)$, we are interested in the standard setting where the batch size is $1$, and the algorithm {does not know} the parameter $\sigmaOne$.} and {cannot} be controlled simply by scaling down the step size. Indeed, this additional bias can be problematic, since the ``positive drift'' could {completely} cancel out the ``negative drift'', i.e., the $-\tetat \gradtsq$ term, in \eqref{eq:standardAnalysisRewritten}. 
Handling this combination of negative and positive drifts constitutes our second challenge.

 \section{Main Results}
\label{sec:convergenceAdaGradNorm}

In this section, we sketch out the key ideas that go into deriving a bound on the convergence rate {\color{cblue} of AdaGrad-Norm }to a first order stationary point. Our main result is the following:

{\color{cblue}
\begin{custtheorem}[Informal statement of \cref{thm:affineVariance}]\label{thm:informalAffineVariance}
{\color{cblue} With probability at least $1-\delta$, t}he iterates of \eqref{eq:alg} satisfy:
\begin{align}\label{eq:mainBodyThmSqrtT}
    \min_{t\in [T]} \gradtsq{} 
    \leq \frac{\constThmMain\log^{\nicefrac{13}{4}}(T)}{\sqrt{\delta^3 T}},
\end{align}
where $\constThmMain \propto (1+\sigmaOne)\parens{\nicefrac{\fzero - \fstar}{\eta} + b_0 + \sigmaZero + (1+\sigmaOne^2)\gradzeronorm + (1+\sigmaOne^6)\eta L}^2 + o(\nicefrac{1}{T})$.\footnote{We use the notation $x \propto y$ to mean $\beta \cdot y \leq x \leq \alpha \cdot y$ for some absolute constant $\alpha > \beta$ independent of all problem parameters.} Moreover, when $\sigmaOne \leq \nicefrac{1}{8}$, then with probability at least $1-\delta$,
\begin{align}\label{eq:mainBodyThmT}
    \min_{t\in [T]} \gradtsq{} 
    \leq \frac{(\sigmaZero A + \sigmaOne B) \log^{\nicefrac{9}{4}}(T)}{\delta^2\sqrt{T}} + \frac{C'\log^{2}(T)}{\delta^2 T},
\end{align}
where $A\propto \nicefrac{\fzero - \fstar}{\eta} + \sigmaZero + \eta L$, $B \propto (1+\sigmaOne^{\nicefrac{3}{2}})(b_0 + \sigmaZero + \gradzeronorm + \eta L {\color{cmaroon}+ \nicefrac{\fzero - \fstar}{\eta})^2}$, and $C' \propto {\color{cmaroon} (1+\sigmaOne^2)(b_0 + \sigmaZero + \eta L + \nicefrac{\fzero - \fstar}{\eta})^2}$.
\end{custtheorem}

\cref{thm:informalAffineVariance} demonstrates two interesting regimes for our guarantee. Namely, \eqref{eq:mainBodyThmSqrtT} shows a $\tO(\nicefrac{1}{\sqrt{T}})$ convergence rate for any choices of $b_0, \eta > 0${\color{cmaroon}, thus establishing our parameter-free guarantee.} However, this bound does not recover the $\nicefrac{1}{T}$ convergence rate in the ``small-noise'' regime. {\color{black} Through a minor modification to the proof technique used to obtain \eqref{eq:mainBodyThmSqrtT}, we are able to derive \eqref{eq:mainBodyThmT}, which demonstrates that \eqref{eq:alg} recovers an $\tO(\nicefrac{1}{T})$ rate of convergence when $\sigmaZero,\sigmaOne = \O(\nicefrac{1}{\sqrt{T}})$ -- the rate obtainable by a well-tuned gradient descent in the noiseless regime up to logarithmic factors. We emphasize that \eqref{eq:alg} does not require a priori knowledge of the smoothness parameter $L$ or the variance parameters $\sigmaZero, \sigmaOne$ to obtain either of the convergence rates in \eqref{eq:mainBodyThmSqrtT} or \eqref{eq:mainBodyThmT}. Indeed, \eqref{eq:alg} adapts automatically to obtain the faster rate in the ``small-noise'' regime.} 

}

As highlighted in \cref{sec:challenges}, obtaining \cref{thm:informalAffineVariance} has two main obstacles: (1) devising a way to deal with the additional $\bias$ term introduced by the affine variance scaling, and (2) lower bounding the step size proxy (for which, as we discussed, it suffices to upper bound $\EXP{\sum_{t\in[T]} \gradtsq}$). We now outline the main ideas needed to overcome each of these.

\subsection{Bounding the Bias via a Compensation Argument}
As displayed in \eqref{eq:adaptiveAnalysisAffineBias}, the affine variance scaling introduces additional bias that our analysis must handle. Indeed, this bound taken together with \eqref{eq:standardAnalysisRewritten} implies the following lemma.
\begin{restatable}{custlemma}{restateAnalogueAffine}
\label{lem:analogueAffine}
Let us recall the step size proxy, $\tetat$, from \cref{def:stepSizeProxy}.
Then, we have that
\begin{align*}
    \frac{\tetat}{2}\parens{1-\sigmaOne\ \! \bias}\gradtsq{}
    \leq \Et{\ft{} - \ftplus}+ \cCommon\ \! \Et{\frac{\sgradtsq}{b_{t-1}^2 + \sgradtsq{}}},
\end{align*}
where $\bias := 4\sqrt{\Et{\nicefrac{\sgradtsq{}}{(b_{t-1}^2 + \sgradtsq{})}}}$ is the additional bias term introduced by the affine variance scaling and $\cCommon=2\sigmaZero\eta + \nicefrac{L \eta^2}{2}$.
\end{restatable}
By \cref{lem:analogueAffine}, whenever $\bias \geq \nicefrac{1}{\sigmaOne}$, we cannot upper bound $\tetat \gradtsq$ as we could in the bounded variance case ($\sigmaOne=0$). To overcome this issue, we utilize the following new ideas.

\paragraph{{Key Idea: Focus on the ``good'' times}.}
Note that, as long as $\bias$ is small, the bound in \cref{lem:analogueAffine} is still useful. Hence, instead of summing both sides of the expression in \cref{lem:analogueAffine} for all times $t\in [T]$, we need to focus on the good events and separate them from the bad events in which $\bias > \nicefrac{1}{\sigmaOne}$. To do so, we first formally define the good time instances as follows.
\begin{custdefinition}[``Good'' times]\label{def:goodTimes}
Using the notation from \cref{lem:analogueAffine}, we call a time $t\in [T]$ ``good'' if $ 1 - \sigmaOne \bias \geq \frac{1}{2},$
and denote $\Sg$ as the set of all such times in the interval $[T]$. Similarly, we call a time $t\in [T]$ ``bad'' if it is not ``good,'' and take $\Sg^c$ as the set of all bad times. 
\end{custdefinition}
By this definition, the ``good'' times are those for which a bound on $\tetat\gradtsq$ is preserved. By summing the expression in \cref{lem:analogueAffine} over only the ``good'' times and applying the second inequality from \eqref{eq:stepSizeControl}, we can derive the following result.
\begin{custlemma}[Informal statement of \cref{lem:summedBoundGoodSet}]\label{lem:informalSummedBoundGoodSet}
Recall the step size proxy of \cref{def:stepSizeProxy} and the notation in \cref{def:goodTimes}. With $\cCommon=2\sigmaZero\eta + \nicefrac{L \eta^2}{2}$, we obtain
\begin{equation}\label{eq:good_events_sum}
    \EXP{\sum_{t\in \Sg}\!\! \frac{\tetat}{4} \gradtsq{}\!}
\leq \fzero - \fstar 
    + \cCommon \log(\poly(T))
    +\EXP{\sum_{t\not\in \Sg} \!\!\ftplus{} - \ft}\!,
\end{equation}
\end{custlemma}
The above expression is {almost} the same as the expression \eqref{eq:standardAnalysisRewrittenExpSummed} which was obtainable in the bounded variance case. The main differences are: (i) the residual term involving the deviations at the ``bad'' times, and (ii) the summation over only $\Sg$ instead of all times $[T]$. Since most times are typically ``good'', as we show in \cref{lem:informalGoodSet}, (ii) is not a serious issue. However, the magnitude of the deviations in ``bad'' times could be large, casting (i) a more serious hurdle. 

\begin{custlemma}[Informal statement of \cref{lem:goodSet}]\label{lem:informalGoodSet}
    Let $\Sg$ be the set of ``good'' times from \cref{def:goodTimes}.
    Then, we have that{\color{cblue}, when $\sigmaOne \leq \nicefrac{1}{8}$, then $|\Sg^c| = 0$, and otherwise}\footnote{As an aside, using essentially the same arguments, we can show that $|\Sg^c|$ satisfies the Bernstein condition with parameter $\const \cdot \log(T)$, which implies that, with high probability, $|\Sg^c| \leq \const\cdot \log^2(T)$.}
    \begin{align*}
        \EXP{|\Sg^c|} \leq 64\sigmaOne^2\log(\poly (T))
        \quad \text{and} \quad
        \EXP{|\Sg^c|^2} \leq \parens{64\sigmaOne^2(1 + 128 \sigmaOne^2) + 2}\log^2(\poly (T)).
    \end{align*}
\end{custlemma}
\begin{proofsketch}
An alternative condition that is equivalent to the one in  \cref{def:goodTimes} is $t$ is ``good'' if $\Et{\etat^2\sgradtsq} \leq \frac{\eta^2}{64\sigmaOne^2}.$
{\color{cblue} Since $\etat^2\sgradtsq \leq \eta^2$ by construction of \eqref{eq:alg}, it follows immediately that all times are ``good'' (i.e., $\Sg^c = \emptyset$) whenever $\sigmaOne \leq \nicefrac{1}{8}$. In the opposite case, this alternate condition}
allows us to argue about the expected number of ``bad'' times via a pigeonholing argument. Specifically, by the tower rule of expectations and the definition of $\etat$, one can show (see \cref{cor:logBound} for details) that
\begin{align*}
\EXP{\sum_{t\in[T]} \Et{\etat^2 \sgradtsq}}
=\EXP{\sum_{t\in[T]} \etat^2 \sgradtsq} 
=\eta^2 \EXP{\log(\nicefrac{b_{T}^2}{b_0^2})} 
= \eta^2 \log(\poly(T)).
\end{align*}
Hence, if more than $64\sigmaOne^2 \log(\poly(T))$ times were ``bad'' in expectation, then since each bad time leads to $\Et{\etat^2\sgradtsq} > \nicefrac{\eta^2}{64\sigmaOne^2}$, we would reach a contradiction to the above bound. 
\end{proofsketch}

This result shows that most times are  ``good.'' Hence, replacing the sum over all time instances with summation over good time instances in \eqref{eq:good_events_sum} would not be a major issue, as long as we can ensure that the additional term corresponding to the bad events, i.e., $\mathbb{E}[\sum_{t\not\in \Sg} \ftplus{} - \ft]$, would not lead to a vacuous upper bound. Next, we formally show how this goal can be achieved.

\vspace{-1mm}
\paragraph{{Key Idea: Compensating for the ``bad'' times}.} \cref{lem:informalSummedBoundGoodSet} shows that, even when we focus on the good times, we still must argue about the deviations at bad times to obtain a convergence guarantee. In order to address this problem, we begin by rewriting \cref{lem:informalSummedBoundGoodSet} by: (i) upper bounding the ``bad'' times using the (potentially quite large) bound obtained from \cref{lem:analogueAffine}, and (ii) subtracting some of the ``good'' deviation terms from both sides to compensate for the bad terms.

Henceforth, we associate each ``bad'' time $t$ with a set of compensating ``good'' times, denoted by $\Scomp{t}$, such that all compensating sets are disjoint. Further, we denote the union of these sets with $\Sc := \cup_{t\in\Sg^c} \Scomp{t}$ and the remaining good time steps with $\tSg := \Sg\setminus \Sc$.
Hence, immediately from \cref{lem:informalSummedBoundGoodSet}, we derive the following.

\begin{custlemma}[Informal statement of \cref{lem:summedBoundModifiedGoodSet}]\label{lem:informalSummedBoundModifiedGoodSet}
In the same setting as \cref{lem:informalSummedBoundGoodSet}, we have that
\begin{align*}
    \EXP{\sum_{t\in \tSg}  \frac{\tetat}{4}\gradtsq{}}
    &\leq \fzero - \fstar
    + \cCommon\log(\poly(T))\\
    &\quad\!\!+ \EXP{\sum_{t\not\in \Sg}\!\!\! \parens{\frac{(4\sigmaOne \!-\! 1)}{2} \tetat \!\gradtsq
    -\!\! \!\sum_{t'\in \Scomp{t}} \frac{\tetaat{t'}}{4} \gradsq{t'}}},
\end{align*}
where $\tSg := \Sg \setminus \Sc$ are remaining ``good'' times after compensation, and $\cCommon=2\sigmaZero\eta + \nicefrac{L \eta^2}{2}$.
\end{custlemma}
The above expression is promising in the following sense. 
If for every ``bad'' time $t \in \Sg^c$ one could find enough compensating ``good'' times $t'\in\Scomp{t}$ with $\tetaat{t'} \gradsq{t'}$ of the same order as the analogous term for $t$, then the last term in \cref{lem:informalSummedBoundModifiedGoodSet} could be bounded {deterministically} as a function of the size of the bad set, $|\Sg^c|$. By \cref{lem:informalGoodSet}, both the size of this set and its square are no more than $\O(\poly\log(T))$ in expectation. Hence, this bound suffices to recover an  expression similar to \eqref{eq:standardAnalysisRewrittenExpSummed}.
The next lemma gives insight into how one can select such ``compentating'' times.

\begin{restatable}{custlemma}{restateCompensationInsight}\label{lem:compensationInsight}
\hspace{-0.05in} Recall the step size proxy $\tetat$ from \cref{def:stepSizeProxy}.
For any time $t\in[T]$ and set $\Scomp{t}\hspace{-0.05in} \subset [T]$ such that (i) $t > \max(\Scomp{t})$ and (ii) $|\Scomp{t}| = \ncomp := {\color{cblue}\max\{8\ceil{4\sigmaOne-1},0\}}$, 
\begin{align*}
   \frac{4\sigmaOne-1}{2}\tetaat{t}\gradsq{t} - \sum_{t'\in \Scomp{t}}\frac{\tetaat{t'}}{4}\gradsq{t'}
   &\leq
   \frac{\eta^2 L \ncomp}{8} (t-\min(\Scomp{t})).
\end{align*}
\end{restatable}

\begin{figure}[t]
    \centering
\includegraphics[scale=0.8]{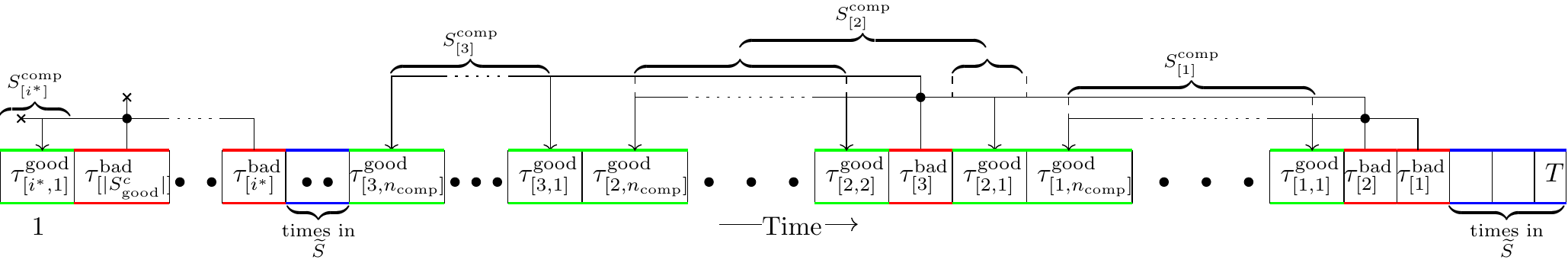}
    \caption{A possible configuration of each bad time $\tb{i}\in\Sg^c$ and the associated \emph{compensating} good times $\Scomp{i}$ from the \emph{greedy} construction in \cref{lem:compSetExists} on the interval $[T]$. Observe that, by this greedy construction, $\tb{1}$ has the \emph{largest} $\ncomp$ ``good'' times in its compensation set, $\Scomp{1}$. The remaining compensation sets are built \emph{greedily} from the \emph{largest} time to the \emph{smallest}. Hence, $\tb{i^*}$ has only a \emph{single} compensating time, and all smaller bad times have \emph{no} compensating times. Finally, note that the number of ``bad'' times, $|\Sg^c|$, is typically quite small relative to $T$ (see \cref{lem:goodSet}), even though it is not depicted as such in the above figure. 
}

\label{fig:compensationExample}
\end{figure}

The above result serves as our guide for constructing the set $\Sc$ to upper-bound the residual term from \cref{lem:informalSummedBoundModifiedGoodSet}. Indeed, it tells us that, in order to bound the deviation $\tetat\gradtsq$ at a bad time $t$, we should not simply pick arbitrary ``good'' times to offset this deviation. Instead, we should pick times which are as close as possible (in time) to $t$. Perhaps surprisingly, it suffices to use the deviations at a {constant} $\ncomp = \O(\sigmaOne)$ number of ``good'' times to compensate for the deviation of $t$. Importantly, these ``good'' times which we choose must come {earlier} in time than $t$, since the step size proxies are ``effectively'' decreasing over time, and thus the proxies at good times after $t$ might be significantly smaller than $\tetat$. To see why selecting these nearby earlier times suffice, recall that the AdaGrad-Norm algorithm \eqref{eq:alg} always takes steps of constant length. Therefore, by $L$-smoothness, the gradients at nearby time steps must also be of the same order. Thus, by choosing nearby, earlier compensating times, we can ensure that both (i) the gradients and (ii) the step size proxies at these good times are of the same order as those of the bad time. {\color{cmaroon}We describe this construction in full detail in \cref{sec:appendix:compensation}, where we additionally include \cref{fig:compensationExample}, which shows an example configuration of these compensating ``good'' times.}

This greedy compensation construction alone, however, is not sufficient to bound the residual term in \cref{lem:informalSummedBoundModifiedGoodSet}, since it might be the case that some bad time $t\in\Sg^c$ has insufficiently many good times to compensate for it (e.g., if $t=1$). We show in \cref{lem:compSetExists} that, whenever we cannot compensate for $t$, this time must, in fact, be very small ($\O(\log(T))$ in expectation). Moreover, since the deviation at any time $t$ can be upper bounded by $t$ (because $\tetat\gradtsq \leq \eta\gradtnorm = \O(t)$ by \cref{lem:polyGradBound}), the corresponding deviation must also be $\O(\log(T))$ in expectation. Further, our greedy construction deterministically guarantees (as we show in \cref{lem:compSetExists}) that compensating times for $t$ will never be more than $\O(|\Sg^c|)$ time steps away from $t$. Thus, the deviations at these bad times also will never be more than $\O(\log(T))$ in expectation, by \cref{lem:compensationInsight,lem:polyGradBound}.

\begin{restatable}{custlemma}{restateCompSetExists}\label{lem:compSetExists}
There exists a construction of $\Sc = \cup_{t \in [\Sg^c]} \Scomp{t}$, where $\Scomp{t}$ denotes the compensating ``good'' times for a bad time $t\in\Sg^c$ (disjoint from other $\Scomp{\tilde{t}}$), satisfying $|\Scomp{t}|\!\leq\! \ncomp\!:=\!{\color{cblue}\max\{8\ceil{4\sigmaOne\!-\!1},0\}}$ and $t \!>\! \max(\Scomp{t})$, where one of the these holds:
\begin{enumerate}
    \item $|\Scomp{t}| = \ncomp$ and{\color{cblue}, if $\ncomp>0$, then} $t - \min(\Scomp{t}) \leq \ncomp\cdot |\Sg^c|$
    \item $|\Scomp{t}| < \ncomp$ and $t\leq \ncomp\cdot |\Sg^c|$
\end{enumerate}
\end{restatable}

\begin{figure}[t]
    \centering
\includegraphics[scale=0.8]{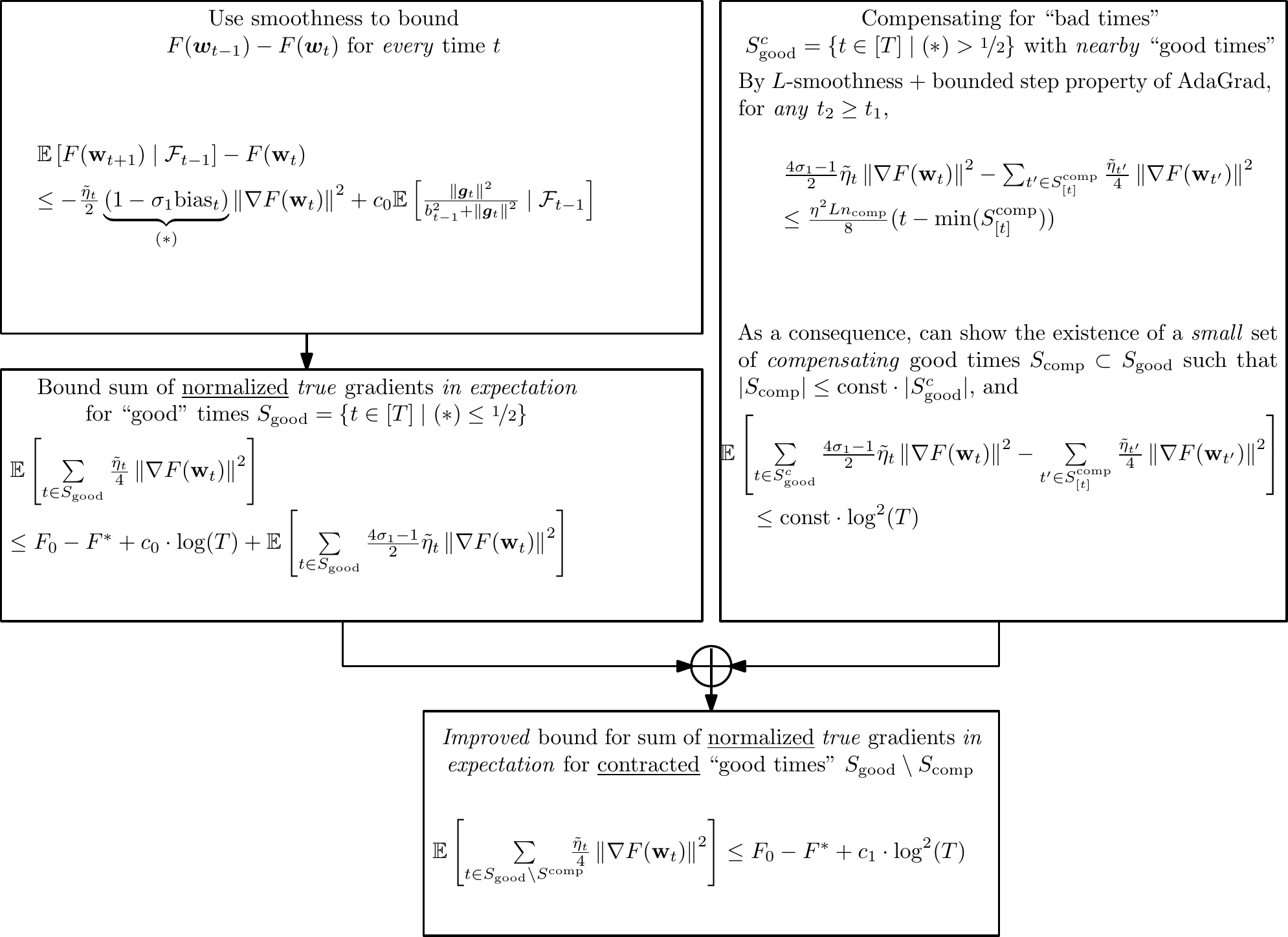}
    \caption{A flow chart of the main ideas underlying the compensation argument used in \cref{lem:mainSummedBoundAffineInterpreted}}
    \label{fig:affineVarCompensation}
\end{figure}

By condition $1$ of \cref{lem:compSetExists} combined with \cref{lem:compensationInsight}, the deviation at a ``bad'' time $t$ can always be bounded by $\O(|\Sg^c|)$ whenever there are enough times to compensate for it. Whenever there are not enough compensating times for $t$, condition $2$ of \cref{lem:compSetExists} implies that this time $t$, and thus also the associated deviation (as we discussed above), must be bounded by $\O(|\Sg^c|)$. Therefore, the total deviation cannot be more than $\O(|\Sg^c|^2)$, which is $\O(\log^2(T))$ in expectation by \cref{lem:informalGoodSet}. Through these observations, we obtain our desired bound, the analogue of \eqref{eq:standardAnalysisRewrittenExpSummed}. {\color{cmaroon}For more details on the arguments presented here, refer to \cref{sec:appendix:compensation}, where we include all proofs, as well as a flow-chart of the main ideas in \cref{fig:affineVarCompensation}.}

\begin{custlemma}[Informal statement of \cref{lem:mainSummedBoundAffineInterpreted}]\label{lem:informalMainSummedBoundAffineInterpreted}
Let the set $\Sc$ from \cref{lem:informalSummedBoundModifiedGoodSet} be chosen as in \cref{lem:compSetExists}.
Then, denoting $\tSg := \Sg\setminus\Sc$ as the set of ``good'' times after compensation,
\begin{align}\label{invariant_upper_bound}
    \EXP{\sum_{t\in \tSg} \frac{\tetat}{4} \gradtsq{}}
    \leq \fzero - \fstar
    + \constInvariance \cdot \log^2(T),
\end{align}
{\color{cblue} where $\constInvariance \propto \eta\sigmaZero + L\eta^2 + \parens{\eta \sigmaOne^2 \gradzeronorm + L\eta^2 \sigmaOne^6}\1{\sigmaOne > \nicefrac{1}{8}}.$
}
\end{custlemma}

With \cref{lem:informalMainSummedBoundAffineInterpreted} in place, we are very close to obtaining a $\tO(\nicefrac{1}{\sqrt{T}})$ convergence rate. Indeed, if we knew deterministically that $\tetat = \Om(\nicefrac{1}{\sqrt{T}})$,  
then substituting in \cref{lem:informalMainSummedBoundAffineInterpreted}, we could conclude that $\EXP{\sum_{t\in\tSg} \gradtsq} = \O(\sqrt{T}\log^2(T))$. This would immediately imply a convergence rate of $\EXP{\min_{t\in[T]} \gradtsq} = \tO(\nicefrac{1}{\sqrt{T}})$ by lower bounding the average by the minimum, and noting that $|\tSg|=\Om(T)$ with high probability (an easy consequence of \cref{lem:informalGoodSet,lem:compSetExists}). However, since $\tetat$ is a random variable which can be significantly smaller than $\nicefrac{1}{\sqrt{T}}$ on some sample paths, deriving the required bound is challenging. Below, we formally show how we address this.

\subsection{Bounding the Expected Sum of Gradients via Recursive Improvement}

As mentioned above, to finalize our convergence result, we need to show that $\EXP{\tetat} = \Om(\nicefrac{1}{\sqrt{T}})$. However, the naive bound one can derive for $\EXP{\tetat}$ as an immediate corollary of \cref{lem:polyGradBound} is 
worse than what we require. We show instead that we can start with a loose lower bound on $\tetat$ which holds with sufficiently high probability and recursively improve it to obtain our desired bound on $\EXP{\tetat}$. 

\paragraph{Key Idea: Recursively-{\color{black}improving} inequalities.} 
We initialize the recursion with an upper bound on  $\EXP{\sum_{t\in[T]} \gradtsq} \leq \O(T^2\log(T))$ from \cref{lem:polyGradBound}, and use this to derive a lower bound on $\tetat$ with high probability ({\em Step 1}). Next we use the upper bound on the expected sum of products, $\sum \tetat \gradtsq$ (the caveat being that the sum is over most but not all of the time indices), from \cref{lem:informalMainSummedBoundAffineInterpreted} to decrease the upper bound on 
$\EXP{\sum_{t\in[T]} \gradtsq}$ ({\em Step 3}). This iteration is now recursed ad infinitum, resulting in \cref{lem:informalRecursiveInterlacing}. Crucial to this iteration is the observation that the upper bound in \cref{lem:informalMainSummedBoundAffineInterpreted} remains unchanged even as the lower bound on $\tetat$ and upper bound on  $\EXP{\sum_{t\in[T]} \gradtsq}$ evolve -- hence, we term \cref{lem:informalMainSummedBoundAffineInterpreted} as the ``invariant upper bound'' property.
{\color{black} While this description gives the main intuition, using this requires more care (see {\em Steps 2 and 3}) because the relation between $\tetat$ and $\sum_{t\in[T]} \gradtsq$ is over all times, whereas the upper bound in  \cref{lem:informalMainSummedBoundAffineInterpreted} contains only the ``good'' times that are not used for compensation.}

\textit{Step 1: Lower bounding $\tetat$.} We start with an upper bound on the expected sum of gradients, $\EXP{\sum_{t\in[T]} \gradtsq} \leq \constRecImp T^{x}\log^y(h(T))$, where $\constRecImp$ is a sufficiently large constant, $h(T)$ is a polynomial function of $T$, and $x$ and $y$ are parameters which can initially, as a consequence of \cref{lem:polyGradBound}, be chosen as $x=2$ and $y=1$. This directly implies an analogous bound on $\EXP{b_{T-1}^2}$ (recall that $b_t$ is defined in \eqref{eq:alg}) through \eqref{eq:sgradBound}. Thus, one immediately obtains, through Markov's inequality, a loose upper bound on $b_{T-1}^2 \leq \constRecImp\ T^{x+\gamma_1}\log^{\gamma_2}(h(T))$ which holds with probability at least $1-\O(\nicefrac{\log^{y-\gamma_2}(h(T))}{T^{\gamma_1}})$ (where we set {\color{cblue}$\gamma_1 = \nicefrac{(4-x)}{3}$} and {\color{cblue}$\gamma_2 = \nicefrac{2(y-1)}{3}$}). Thus, taking $\E_T(\delta)$ to be this high probability event, and applying the deterministic bound on $\gradtnorm$ from \cref{lem:polyGradBound}, we obtain a lower bound for each $\tetat$ whenever $\E_T(\delta)$ is true, which we use to obtain:
\begin{equation}\label{eq:interlacing1_re_main}
    \EXP{\sum_{t\in\tSg} \tetat \gradtsq}
   \! \geq \EXP{\sum_{t\in\tSg} \tetat \gradtsq \1{\E_T(\delta)}}\!\geq \frac{\eta\EXP{\sum\limits_{t\in\tSg} \gradtsq \1{\E_T(\delta)}}}{\sqrt{2 \constRecImp\ T^{x+\gamma_1}\log^{\gamma_2}(h(T))}}.
\end{equation}

\textit{Step 2: Bounding the ``good'' terms.} To remove the indicator function in the lower bound, one can use the fact that
   $\EXP{\sum_{t\in\tSg} \gradtsq \1{\E_T(\delta)}} = \EXP{\sum_{t\in\tSg} \gradtsq (1-\1{\E_T(\delta)^c})},$
and the polynomial upper bound that we have on the gradients sum from \cref{lem:polyGradBound} together with an upper bound on the failure probability of $\E_T(\delta)^c$. Moreover, we importantly use the ``invariant'' upper bound on $\EXP{\sum_{t\in\tSg}\tetat\gradtsq}$ from \cref{lem:informalMainSummedBoundAffineInterpreted} together with the lower bound on this same quantity from \eqref{eq:interlacing1_re_main} to conclude that 
\begin{align}\label{eq:interlacing3_main}
    \EXP{\sum_{t\in\tSg} \gradtsq}
    &\leq 
    \frac{\constRecImp}{2} T^{\frac{x + 2}{3}}\log^{\frac{y + 5}{3}}(h(T)).
\end{align}
Note that this is {almost} an improved bound on $\EXP{\sum_{t\in[T]} \gradtsq}$. However, the summation range in \eqref{eq:interlacing3_main} is a subset of $[T]$ that almost has the same size.

\begin{figure}
    \centering
\includegraphics[scale=0.8]{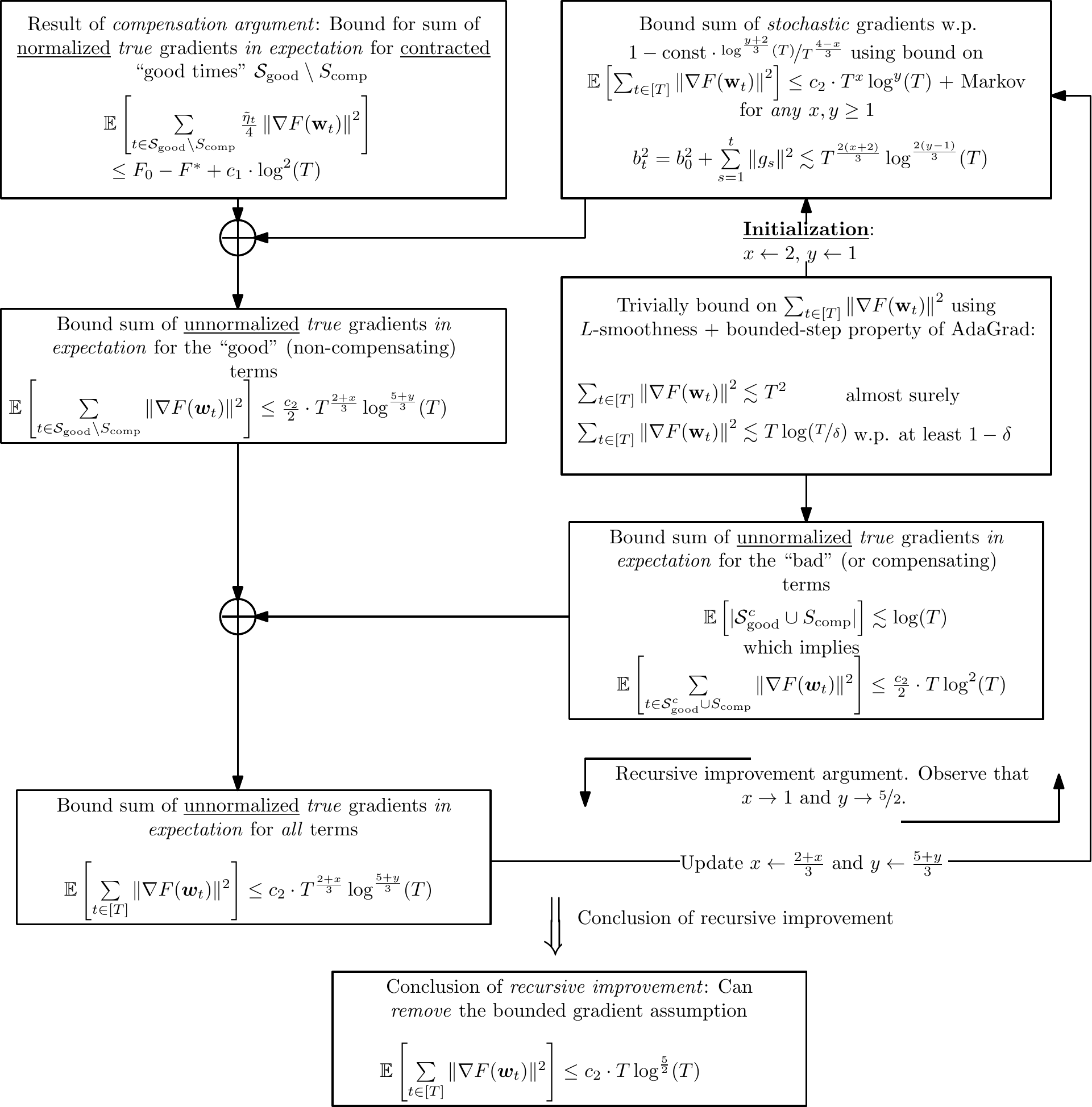}
    \caption{A flow chart of the main ideas underlying the ``Recursive Improvement'' argument of \cref{lem:recursiveInterlacing}. }
    \label{fig:affineVarChickenEgg}
\end{figure}

\textit{Step 3: Bounding the ``bad'' terms.}
It remains only to bound $\EXP{\sum_{t\not\in\tSg} \gradtsq}$. Recall that, by construction of $\Sc$ in \cref{lem:compSetExists}, $|\Sc| \leq \ncomp \cdot |\Sg^c|$. Further, by the result in \cref{lem:polyGradBound}, each $\gradtsq = \O(T\log(T))$ with probability at least $1-{T^{-2}}$, and $\O(T^2)$ deterministically. Hence, by using \cref{lem:informalGoodSet} to bound the expected size of $|\Sg^c|$, we obtain that
\begin{align}\label{eq:interlacing4_main}
    \EXP{\sum_{t\not\in\tSg} \gradtsq}
    &\leq 
    \frac{\constRecImp}{2} {\color{cblue}T\log^{2}(h(T))}.
\end{align}
Thus, by combining the results of \eqref{eq:interlacing3_main} and \eqref{eq:interlacing4_main} (recalling the constraint that $x\geq 1$), we conclude that $\EXP{\sum_{t\in[T]}\gradtsq} \leq \constRecImp\ T^{\nicefrac{(x+2)}{3}}\log^{\nicefrac{(y+5)}{3}}(h(T))$. We may thus use this improved bound recursively in place of the original choice of $x$ and $y$ from \textit{Step 1}.  The conclusion of this ``recursive improvement'' argument is that $\EXP{\sum_{t\in[T]} \gradtsq} = \tO(T)$, which, by Jensen's inequality, implies  $\EXP{\tetat} = \tOm(\nicefrac{1}{\sqrt{T}})$. This result is summarized in \cref{lem:informalRecursiveInterlacing}. {\color{cmaroon}For more details on the arguments presented here, refer to \cref{sec:appendix:recursiveInterlacing}, where we include all proofs, as well as a flow-chart of the main ideas in \cref{fig:affineVarChickenEgg}.}

\begin{custlemma}[Informal statement of \cref{lem:recursiveInterlacing}]\label{lem:informalRecursiveInterlacing}
Suppose that, for some parameters $x \in [1,4]$, $y \geq 1$, $h(T)$ a {\color{cblue}sufficiently large} polynomial function of $T$, and sufficiently large constant $\constRecImp$,
{\color{cblue}
\begin{align*}
    \constRecImp \propto b_0^2 + \sigmaZero^2 + \max\braces{1,\sigmaOne^{12}}(\gradzerosq + \eta^2 L^2) + \parens{\nicefrac{\fzero - \fstar}{\eta}}^2,
\end{align*}
}
we have that
$\EXP{\sum_{t\in[T]} \gradtsq} \leq \constRecImp\ T^x \log^y(h(T))$.
Then, the following tighter bound also holds:
\begin{align}\label{eq:interlacingResult1_main}
    \EXP{\sum_{t\in [T]} \gradtsq}
    &\leq 
    \constRecImp T^{\frac{x + 2}{3}}\log^{\frac{y + 5}{3}}(h(T)).
\end{align}
In particular, as a consequence of \cref{lem:polyGradBound},
\begin{align}\label{eq:interlacingResult2_main}
    \EXP{\sum_{t\in[T]} \gradtsq} \leq \constRecImp\ T \log^{\frac{5}{2}}(h(T))
    \quad \text{and} \quad
    \EXP{\tetat} = \tOm(\nicefrac{1}{\sqrt{T}}).
\end{align}
\end{custlemma}

\subsection{Wrapping up}

With these {\color{cblue}bounds from \cref{lem:informalMainSummedBoundAffineInterpreted,lem:informalRecursiveInterlacing}} in place, obtaining the convergence result for {\color{cblue}\eqref{eq:alg} in \cref{thm:informalAffineVariance}} is immediate. Indeed, we note that \cref{lem:informalMainSummedBoundAffineInterpreted} gives us essentially the same bound 
as the one obtainable in the uniformly-bounded variance case \eqref{eq:standardAnalysisRewrittenExpSummed} (modulo the summation over the set $\tSg$ instead of all times $[T]$). Therefore, we may apply (essentially) the same H\"older's inequality argument as in \citep{WWB19}, replacing their application of the uniform gradient bound with {\color{cblue} our bound on the expected sum of gradients from \cref{lem:informalRecursiveInterlacing}}, and taking extra care that our summation from \cref{lem:informalMainSummedBoundAffineInterpreted} is over a {random} set $\tSg$. 
We give the full proof of this theorem in \cref{sec:appendix:convergenceAdaGradNorm}.

\begin{custremark}\label{ada_grad_remark}
While we focus in this paper on the convergence rate of one particular adaptive SGD method, our methods are not overly specialized to AdaGrad-Norm. Indeed, using nearly identical arguments per coordinate, we can obtain similar $\tO(\nicefrac{1}{\sqrt{T}})$ convergence rates under similar assumptions for coordinate-wise AdaGrad, albeit with an additional polynomial dependence on $d$.
\end{custremark} \section{Conclusion}
\label{sec:concl}

In this paper, we extended the analysis of AdaGrad-Norm to the setting where the gradients are possibly unbounded and the noise variance scales affinely. We showed that under these conditions, together with the standard smoothness assumption, the iterates of \eqref{eq:alg} reach a first-order stationary point of a nonconvex function with an error of $\O(\nicefrac{\poly\log(T)}{\sqrt{T}})$.

\section*{Acknowledgements}
This research is supported in part by NSF Grants 1952735, 1934932, 2019844, 2127697, and 2112471, ARO Grant W911NF2110226, AFOSR MURI FA9550-19-1-0005, the Machine Learning Lab (MLL) at UT Austin, and the Wireless Networking and Communications Group (WNCG) Industrial Affiliates Program.

\printbibliography

\newpage

\appendix

\section{Preliminaries}
\label{sec:appendix:preliminaries}
Here, we provide proofs for claims from \cref{sec:preliminaries}, as well as some auxiliary results and notation. We additionally state some definitions that will be useful for proving our results.

\begin{restatable}{custlemma}{restateLogBound}\label{lem:logBound}
For any sequence $\{a_s\}_{s=0}^\infty$ such that $a_0 > 0$ and $a_s\geq 0$ for all $s$,
\begin{align*}
    \sum_{t=0}^T \frac{a_t}{\sum_{s=0}^t a_s}
    \leq 1 + \log\parens{\sum_{t=0}^T a_t} - \log\parens{a_0}
\end{align*}
\end{restatable}

\begin{custproof}
The base case of $T=0$ holds with equality. Let us now assume that the claim holds at $T$. Then, we have that
\begin{align*}
    \sum_{t=0}^{T+1} \frac{a_t}{\sum_{s=0}^t a_s}
    &\leq
    1 + \log\parens{\sum_{t=0}^T a_t} - \log(a_0) + \frac{a_{T+1}}{\sum_{s=0}^{T+1} a_s}\\
    &\leq
    1 + \log\parens{\sum_{t=0}^T a_t} - \log(a_0) + \log\parens{\frac{\sum_{s=0}^{T+1} a_s}{\sum_{s=0}^T a_s}}\\
    &=
    1 + \logp{\sum_{t=0}^{T+1} a_t} - \log(a_0),
\end{align*}
where the first inequality holds by the induction hypothesis, and the second because of the fact $x < - \log(1-x)$ (where $\log(\cdot)$ denotes the natural logarithm).
\end{custproof}

Our analysis will focus on adaptive gradient algorithms with a particularly convenient structure, which we refer to as the \emph{Bounded Step-Size Property}
\begin{custdefinition}[$\stepBnd$-Bounded Step-Size Property]\label{def:boundedStep}
We say that an optimization algorithm has \emph{$\stepBnd$-Bounded Step-Sizes} if, for \emph{any} pair of adjacent iterates $(\w_t,\w_{t+1})$ generated by the algorithm, the following inequality holds deterministically:
\begin{align*}
    \norm{\w_{t+1} - \w_t} \leq \stepBnd.
\end{align*}
\end{custdefinition}

Another convenient property of the algorithms we study is what we call the \emph{Decay Property}:
\begin{custdefinition}[$(\decayBnd,b_0)$-Decay Property]\label{def:decayProperty}
We say that an optimization algorithm satisfies the \emph{$(\decayBnd,b_0)$-Decay Property} if the iterate sequence $\{\w_t\}_{t\in [T]}$ satisfies the following inequality deterministically:
\begin{align*}
    \sum_{t=1}^T \normSq{\w_{t+1} - \w_t}
    \leq \decayBnd \cdot \logp{1 + \sum_{t=1}^T \frac{\sgradtsq}{b_0^2}}.
\end{align*}
\end{custdefinition}

We observe that these property is satisfied by a number of interesting adaptive gradient algorithms.

\begin{custobservation}\label{obs:boundedStep}
AdaGrad-Norm has \emph{$\eta$-Bounded Step-Sizes} and \emph{$(\eta^2,b_0)$-Decay}. The first follows since for any time $t\geq 0$,
\begin{align*}
    \norm{\w_{t+1} - \w_t} = \eta\frac{\sgradtnorm{}}{\sqrt{b_{t-1}^2 + \sgradtsq{}}} \leq \eta.
\end{align*}
The second is an immediate consequence of \cref{lem:logBound}{\color{cblue}, taking $a_0 = b_0^2$ and $a_s = \sgradsq{s}$ for $s>0$.}
\end{custobservation}
\begin{custobservation}\label{obs:boundedStepCoordinateWise}
Coordinate-wise AdaGrad (with coordinate-dependent step sizes 
\begin{align*}
    \itetat := \frac{\eta}{\sqrt{b_{t-1,i}^2 + \isgradtsq}}   
\end{align*}
has \emph{$\eta \cdot \sqrt{d}$-Bounded Step-Sizes} and \emph{$(d\eta^2,b_0)$-Decay}. The first follows since since $\abs{\w_{t+1,i} - \w_{t,i}} \leq \eta$ for every coordinate $i \in [d]$. The second follows by applying \cref{lem:logBound} to the sum of $|\w_{t+1,i}-\w_{t,i}|^2 = \eta^2 \frac{\isgradtsq}{b_{t-1,i}^2 + \isgradtsq}$ for each coordinate.
\end{custobservation}

{\color{cblue}
\begin{custremark}
We note here that all of the remaining results in this section could be stated in more generality by using \cref{def:boundedStep,def:decayProperty}. To showcase our ideas in the simplest manner, we will state everything in the context of the AdaGrad-Norm algorithm \eqref{eq:alg}.
\end{custremark}
}

By \cref{assump:smooth} and {\color{cblue}\cref{obs:boundedStep}}, we also have the following simple, but quite useful, facts, which give us crude but, crucially, \emph{polynomial} (in $T$) bound on $\gradtsq$:
\begin{restatable}{custlemma}{restateTrivialGradBound}\label{lem:trivialGradBound}
Consider the AdaGrad-Norm algorithm \eqref{eq:alg} running on an $L$-smooth objective function $F$. Then, for any times $t_2 \geq t_1$,
\begin{align*}
    \abs{\gradnorm{t_2} - \gradnorm{t_1}}
\leq {\color{cblue}\eta} L (t_2 - t_1).
\end{align*}
In particular, this implies that
\begin{align*}
\gradtnorm \leq \norm{\gradzero} + {\color{cblue}\eta} L t
\end{align*}
\end{restatable}
\begin{custproof}
The proof follows by first applying the triangle inequality and using a telescoping sum to bound
\begin{align*}
    \abs{\gradnorm{t_2} - \gradnorm{t_1}}
    \leq \norm{\grad{t_2} - \grad{t_1}}
    = \norm{\sum_{s=t_1}^{t_2-1} \grad{s+1} - \grad{s}},
\end{align*}
then noting that, for each $s\in [t_1,t_2]$, by {\color{cblue}\cref{assump:smooth,obs:boundedStep}},
\begin{align*}
    \norm{\grad{s+1} - \grad{s}} \leq L \norm{\w_{s+1} - \w_s} \leq L \cdot {\color{cblue}\eta}.
\end{align*}
\end{custproof}

The above bound on $\gradtnorm$ is quite useful, since it guarantees a \emph{polynomial} (in $T$) bound for $\gradtnorm$. However, note that this bound is \emph{much} more crude than the bound assumed by \citet{WWB19,DBBU20} (where they assumed $\gradtsq \leq B < \infty$ for every $t$). It turns out that, on ``nice'' sample paths, a significantly tighter bound can be derived. Intuitively, these sample paths are those for which the quantity $b_T^2 = b_0^2 + \sum_{t=1}^T \sgradtsq$ is bounded by a polynomial in $T.$
\begin{custdefinition}[Nice event]\label{def:niceEvents}
For any time $s \in {\color{cblue}\{0\}\cup}[T]$ and failure probability $\delta \in (0,1]$, we define the following ``nice event'':
    \begin{align}\label{eq:logBoundEvent}
        \E_s(\delta)
        = 
        \braces{b_s^2 \leq b_0^2 + \frac{s \sigmaZero^2 + (1+\sigmaOne^2) \EXP{\sum_{t\in [s]} \gradtsq}}{\delta}},
    \end{align}
    {\color{cblue}and take $\E_s(\delta) = \emptyset$ for $\delta > 1$.}

    We note that, by construction, Markov's inequality tells us that this event occurs with probability at least $1-\delta$, i.e., $\PRO{\E_s(\delta)^c} \leq \delta$.
    Further, taking
    \begin{align}\label{eq:polyFn}
        f(s) = {\color{cblue}e} + \frac{\sigmaZero^2 s}{b_0^2} + \frac{(1+\sigmaOne^2 )s}{b_0^2}(\gradzeronorm + {\color{cblue}\eta} L s)^2,
    \end{align}
it follows (by upper bounding $\EXP{\sum_{t\in[s]} \gradsq{s}}\leq s(\gradzeronorm + {\color{cblue}\eta} L s)^2$ by \cref{lem:trivialGradBound}) that, whenever $\E_s(\delta)$ is true, we have that $\nicefrac{b_s^2}{b_0^2} \leq \nicefrac{f(s)}{\delta}.$
\end{custdefinition}

As we will soon see, bounding the quantity $\sum_{t\in[T]} \normSq{\w_{t+1}-\w_t}$ will be crucial in many parts of our analysis. Under the ``nice'' events from \cref{def:niceEvents}, this quantity can be easily controlled:
\begin{restatable}{custlemma}{restateLogBoundApplied}\label{cor:logBound}
{\color{cblue}For any choice of $b_0^2$, and any sample path, \eqref{eq:alg} satisfies}
    \begin{align}\label{eq:logBoundDeterministic}
        \sum_{t=1}^T \normSq{\w_{t+1}-\w_t} \leq {\color{cblue}\eta^2}\log\parens{\frac{b_{T}^2}{b_0^2}},
    \end{align}
Further, assuming that the ``nice event'' \eqref{eq:logBoundEvent} ($\E_s(\delta)$) is true at time $s\in[T]$, and taking $f(\cdot)$ as in \eqref{eq:polyFn},
    \begin{align}\label{eq:logBoundHighProbCond}
        \Econd{s}{\sum_{t=1}^T \normSq{\w_{t+1}-\w_t}}
        \leq
        {\color{cblue}\eta^2}\logp{\nicefrac{f(T)}{\delta}}.
    \end{align}
    In particular, since $\E_0(1)$ is {\color{cblue}(trivially)} always true, the above implies that
    \begin{align}\label{eq:logBoundExp}
        \EXP{\sum_{t=1}^T \normSq{\w_{t+1}-\w_t}}
        \leq
        {\color{cblue}\eta^2}\log(f(T)),
    \end{align}
    Additionally, when $\E_T(\delta)$ (the nice event \emph{at time $T$}) is true,
    \begin{align}\label{eq:logBoundHighProbSum}
        \sum_{t=1}^T \normSq{\w_{t+1}-\w_t}
        \leq
        {\color{cblue}\eta^2}\logp{\nicefrac{f(T)}{\delta}}.
    \end{align}
\end{restatable}
\begin{custproof}
We already established \eqref{eq:logBoundDeterministic} in \cref{obs:boundedStep}.
{\color{cmaroon}For the remaining inequalities, we may assume without loss of generality that $\delta \leq 1$. Indeed, whenever $\delta > 1$, then $\E_s(\delta) = \emptyset$ by \cref{def:niceEvents}, and thus $\E_s(\delta)$ is never true, so all of the claims follow trivially.}

To show \eqref{eq:logBoundHighProbCond}, we note that, on any sample path, by \eqref{eq:logBoundDeterministic} and Jensen's inequality,
\begin{align*}
   \Econd{s}{\sum_{t=1}^T \normSq{\w_{t+1}-\w_t}}
    \leq {\color{cblue}\eta^2}\Econd{s}{\logp{\frac{b_T^2}{b_0^2}}}    \leq {\color{cblue}\eta^2}\logp{1+ \sum_{t=1}^s \frac{\sgradtsq}{b_0^2} + \sum_{t=s+1}^T \frac{\Econd{s}{\sgradtsq}}{b_0^2}}.
\end{align*}
To bound this term above, first observe that, as noted in \eqref{eq:sgradBound}, \cref{assump:unbiasedGrad,assump:affineVariance} imply that
\begin{align*}
    \Et{\sgradtsq} 
    \leq \sigmaZero^2 + (1+\sigmaOne^2)\gradtsq.
\end{align*}
Further, when \eqref{eq:logBoundEvent} ($\E_s(\delta)$) is true at time $s$, we have that, by \cref{lem:trivialGradBound},
\begin{align*}
    \frac{1}{b_0^2}\sum_{t=1}^s \sgradtsq
    \leq \frac{s \sigmaZero^2 + (1+\sigmaOne^2) \EXP{\sum_{t\in[s] }\gradtsq}}{b_0^2 \delta}
    \leq \frac{s \sigmaZero^2 + (1+\sigmaOne^2) s \parens{\gradzeronorm + {\color{cblue}\eta} L s}^2}{b_0^2 \delta}.
\end{align*}
Combining the above bounds, we conclude that
\begin{align*}
    \Econd{s}{\sum_{t=1}^T \normSq{\w_{t+1}-\w_t}} 
    \leq {\color{cblue}\eta^2}\logp{1 + \frac{T \sigmaZero^2 + (1+\sigmaOne^2) T\parens{\gradzeronorm + {\color{cblue}\eta} L T}^2}{b_0^2 \delta}} 
    \leq {\color{cblue}\eta^2}\logp{\frac{f(T)}{\delta}},
\end{align*}
as claimed.
Finally, observe that \eqref{eq:logBoundExp} and \eqref{eq:logBoundHighProbSum} follow immediately from \eqref{eq:logBoundHighProbCond}, taking $s=0$ (noting that $\E_0(1)$ is true deterministically) and $s=T$, respectively.
\end{custproof}
With the above construction in place, we are ready to give a \emph{slightly} stronger bound for $\gradtsq$, improving upon \cref{lem:trivialGradBound} (with high probability) in many interesting regimes.
\begin{restatable}{custlemma}{restateLogGradBound}\label{lem:logGradBound}
Consider any time $t\in [T]$ during a run of 
\eqref{eq:alg} initialized at a starting point $\wzero$, and is currently at iterate $\w_t$. Then,
\begin{align*}
    \gradtsq \leq 2 \gradzerosq + 2{\color{cblue}\eta^2} L^2 t \cdot \log\parens{\nicefrac{b_t^2}{b_0^2}}
\end{align*}
and additionally, assuming that $\E_t(\delta)$ from \cref{def:niceEvents} is true, and taking $f(\cdot)$ as in \eqref{eq:polyFn}, then
\begin{align*}
    \gradtsq \leq 2 \gradzerosq + 2{\color{cblue}
    \eta^2} L^2 t \cdot \log\parens{f(t)/\delta}.
\end{align*}
\end{restatable}
\begin{custproof}
The proof follows effectively from the same arguments used to prove \cref{lem:trivialGradBound}, only using the improved bound from \cref{cor:logBound} in place of \cref{lem:trivialGradBound}. Indeed, using the same decomposition, and applying Cauchy-Schwarz, we have that
\begin{align*}
    \gradtsq 
    &\leq 2\gradzerosq + 2 L^2 \parens{\sum_{s=1}^t \norm{\w_{s+1}-\w_s}}^2\\
    &\leq 2\gradzerosq + 2 L^2 t \sum_{s=1}^t \normSq{\w_{t+1}-\w_t}\\
    &\leq 2\gradzerosq + 2 L^2 {\color{cblue}\eta^2} t \log\parens{\frac{b_t^2}{b_0^2}}
\end{align*}
where the first inequality follows from the decomposition used in the proof of \cref{lem:trivialGradBound}, the second follows by Cauchy-Schwarz, and the third from \cref{lem:logBound}.

The second claim follows immediately from the above, combined with \cref{cor:logBound}.
\end{custproof}

 \section{Deriving the Starting Point}
\label{sec:appendix:startingPoint}

Here, we provide the proof for the starting point of our analysis, \cref{lem:analogueAffine}, from \cref{sec:convergenceAdaGradNorm}.

\restateAnalogueAffine*

\begin{custproof}
We will begin by using our assumption of $L$-smoothness, along with the definition of the algorithm, to get the bound:
\begin{align*}
    \ftplus{} - \ft{}
    &\leq \innerProd{\gradt}{\w_{t+1} - \w_t} + \frac{L}{2}\normSq{\w_{t+1} - \w_t}\\
    &= -\etat\innerProd{\gradt}{\sgradt} + \frac{L \etat^2}{2}\sgradtsq\\
    &= -\etat\gradtsq{} - \etat\innerProd{\gradt}{\sgradt{} - \gradt{}} + \frac{L \etat^2}{2}\sgradtsq
\end{align*}
Now, as noted in \cite{WWB19}, the inner product term is \emph{not} zero in expectation, since $\etat$ depends on $\sgradt{}$. Hence, we introduce a step size proxy $\tetat$ from \cref{def:stepSizeProxy}, which \emph{is} independent of $\sgradt{}$ (conditioned on $\F_{t-1}$).
This choice, unlike $\tetat$, satisfies:
\begin{align*}
    \Et{\tetat{} \innerProd{\gradt}{\sgradt{} - \gradt{}}} = 0
\end{align*}
Hence, by taking expectations of our first inequality and adding this mean-zero quantity to the resulting expression, we have that
\begin{align*}
    \Et{\ftplus{}} - \ft{}
    &\leq -\tetat\gradtsq{} - \Et{\parens{\etat{} - \tetat{}}\innerProd{\gradt}{\sgradt{}}}\\
    &\quad+ \Et{\frac{L \etat^2}{2}\sgradtsq}
\end{align*}

We will now focus on bounding the second term. Observe that, denoting
$a=b_{t-1}^2 + \sgradtsq{}$ and $b=b_{t-1}^2 + (1+\sigmaOne^2)\gradtsq{} + \sigmaZero^2$,
\begin{align*}
\abs{\frac{\etat{} - \tetat{}}{\eta}}
= \abs{\frac{1}{\sqrt{a}} - \frac{1}{\sqrt{b}}}
= \abs{\frac{\sqrt{b} - \sqrt{a}}{\sqrt{ab}}}
= \abs{\frac{b - a}{\sqrt{ab}(\sqrt{a}+\sqrt{b})}}
\end{align*}
From this, we conclude that
\begin{align*}
\abs{\frac{\etat{} - \tetat{}}{\eta}}
&= \abs{\frac{(1+\sigmaOne^2)\gradtsq{} + \sigmaZero^2 - \sgradtsq{}}{\sqrt{ab}(\sqrt{a}+\sqrt{b})}}\\
&\leq \frac{\abs{(\gradtnorm{} - \sgradtnorm{})(\gradtnorm{} + \sgradtnorm{})} + \sigmaZero^2 + \sigmaOne^2\gradtsq{}}{\sqrt{ab}(\sqrt{a}+\sqrt{b})}\\
&\leq \frac{\norm{\sgradt{} - \gradt{}} + \sqrt{\sigmaZero^2 + \sigmaOne^2\gradtsq{}}}{\sqrt{b_{t-1}^2 + \sgradtsq{}}\sqrt{b_{t-1}^2 + (1+\sigmaOne^2)\gradtsq{} + \sigmaZero^2}}
\end{align*}

Plugging this bound into the above, and taking expectation with respect to the filtration at $t-1$, we have shown that
\begin{align*}
    &\Et{\ftplus{}} - \ft{}\\
    &\leq -\tetat\gradtsq{}\\
    &\quad+ \eta\Et{\frac{\norm{\sgradt{} - \gradt{}}\gradtnorm{}\sgradtnorm{}}{\sqrt{b_{t-1}^2 + \sgradtsq{}}\sqrt{b_{t-1}^2 + (1+\sigmaOne^2)\gradtsq{} + \sigmaZero^2}}}\\
    &\quad+ \eta\frac{\sqrt{\sigmaZero^2 + \sigmaOne^2\gradtsq{}}\gradtnorm{}}{\sqrt{b_{t-1}^2 + (1+\sigmaOne^2)\gradtsq{} + \sigmaZero^2}} \Et{\frac{\sgradtnorm{}}{\sqrt{b_{t-1}^2 + \sgradtsq{}}}}\\
    &\quad+ \frac{L \eta^2}{2}\Et{\frac{\sgradtsq}{b_{t-1}^2 + \sgradtsq{}}}
\end{align*}
We will now show that the second and third terms above have the same upper bound.
Focus on the second term above, we apply H\"older's inequality and the affine variance assumption to conclude that
\begin{align*}
    &\Et{\frac{\norm{\sgradt{} - \gradt{}}\gradtnorm{}\sgradtnorm{}}{\sqrt{b_{t-1}^2 + \sgradtsq{}}\sqrt{b_{t-1}^2 + (1+\sigmaOne^2)\gradtsq{} + \sigmaZero^2}}}\\
    &\leq
    \sqrt{\frac{\gradtsq{} \Et{\normSq{\sgradt{} - \gradt{}}}}{b_{t-1}^2 + (1+\sigmaOne^2)\gradtsq{} + \sigmaZero^2} \Et{\frac{\sgradtsq{}}{b_{t-1}^2+\sgradtsq{}}}}\\
    &\leq
    \sqrt{\frac{\gradtsq{} \parens{\sigmaZero^2 + \sigmaOne^2\gradtsq{}}}{b_{t-1}^2 + (1+\sigmaOne^2)\gradtsq{} + \sigmaZero^2} \Et{\frac{\sgradtsq{}}{b_{t-1}^2+\sgradtsq{}}}}
\end{align*}
Now, focusing on the third term, by Jensen's inequality to the concave function $\sqrt{\cdot}$, we know that
\begin{align*}
    \Et{\frac{\sgradtnorm}{\sqrt{b_{t-1}^2 + \sgradtsq{}}}}
    =
    \Et{\sqrt{\frac{\sgradtsq}{b_{t-1}^2 + \sgradtsq{}}}}
    \leq
    \sqrt{\Et{\frac{\sgradtsq}{b_{t-1}^2 + \sgradtsq{}}}}
\end{align*}
which show that the second and third terms have exactly the same upper bound. Combining these expressions and rearranging, we find
\begin{align*}
    \Et{\ftplus{}} - \ft{}
    &\leq -\tetat\gradtsq{}\\
    &\quad+ 2\eta\frac{\sqrt{\sigmaZero^2 + \sigmaOne^2\gradtsq{}}\gradtnorm{}}{\sqrt{b_{t-1}^2 + (1+\sigmaOne^2)\gradtsq{} + \sigmaZero^2}} \sqrt{\Et{\frac{\sgradtsq{}}{b_{t-1}^2 + \sgradtsq{}}}}\\
    &\quad+ \frac{L \eta^2}{2}\Et{\frac{\sgradtsq}{b_{t-1}^2 + \sgradtsq{}}}\\
    &\leq -\tetat\gradtsq{}\\
    &\quad+ 2\eta\frac{\sigmaZero\gradtnorm{}}{\sqrt{b_{t-1}^2 + (1+\sigmaOne^2)\gradtsq{} + \sigmaZero^2}} \sqrt{\Et{\frac{\sgradtsq{}}{b_{t-1}^2 + \sgradtsq{}}}}\\
    &\quad+ 2\eta\frac{\sigmaOne\gradtsq{}}{\sqrt{b_{t-1}^2 + (1+\sigmaOne^2)\gradtsq{} + \sigmaZero^2}} \sqrt{\Et{\frac{\sgradtsq{}}{b_{t-1}^2 + \sgradtsq{}}}}\\
    &\quad+ \frac{L \eta^2}{2}\Et{\frac{\sgradtsq}{b_{t-1}^2 + \sgradtsq{}}}
\end{align*}
To conclude, we can bound the second term above, using the inequality $ab \leq \frac{1}{2}a^2 + \frac{1}{2}b^2$, choosing $a=\frac{\sqrt{\eta}\gradtnorm{}}{\sqrt{b_{t-1}^2+(1+\sigmaOne^2)\gradtsq{}+\sigmaZero^2}}$, $b=\frac{2\sqrt{\eta}\sigma}{\sqrt{b_{t-1}^2+(1+\sigmaOne^2)\gradtsq{}+\sigmaZero^2}}\sqrt{\Et{\frac{\sgradtsq{}}{b_{t-1}^2 + \sgradtsq{}}}}$. After grouping the resulting expressions, we arrive at the claimed inequality.
\end{custproof}
 \section{Most Times are (Typically) Good}
\label{sec:appendix:goodTimes}

Here, we provide proofs regarding properties and consequences of the ``good'' times (\cref{def:goodTimes}) from \cref{sec:convergenceAdaGradNorm}.

\begin{custlemma}\label{lem:summedBoundGoodSet}
Recalling the step size proxy of \cref{def:stepSizeProxy} and the notation in \cref{def:goodTimes}, we obtain
\begin{align*}
    \EXP{\sum_{t\in \Sg} \frac{\tetat}{4} \gradtsq{}}
    &\leq \fzero - \fstar 
        + \cCommon \EXP{\sum_{t\in\Sg} \frac{\sgradtsq}{b_{t-1}^2 + \sgradtsq}}
        +\EXP{\sum_{t\not\in \Sg} \ftplus{} - \ft}\\
    &\leq \fzero - \fstar 
        + \cCommon \log(f(T))
        +\EXP{\sum_{t\not\in \Sg} \frac{4\sigmaOne-1}{2} \tetat \gradtsq},
\end{align*}
where $\cCommon = 2\sigmaZero\eta + \nicefrac{L\eta^2}{2}$, and $f(\cdot)$ is the function defined in \eqref{eq:polyFn}.
\end{custlemma}

\begin{custproof}
The proof is an easy consequence of \cref{lem:analogueAffine} together with the fact that $\{t\in \Sg\} \in \F_{t-1}$. Indeed, by construction of $\Sg$, whenever $t\in\Sg$, we have that
\begin{align*}
    1-4\sigmaOne\sqrt{\Et{\frac{\sgradtsq{}}{b_{t-1}^2 + \sgradtsq{}}}}
    \geq \frac{1}{2}.
\end{align*}
Therefore, \cref{lem:analogueAffine} implies that, whenever $t\in\Sg$,
\begin{align}\label{eq:good_event_starting_pt}
    &\Et{\ftplus{} - \ft{}}\nonumber\\
    &\leq
    -\frac{\eta}{4} \frac{\gradtsq{}}{\sqrt{b_{t-1}^2 + (1+\sigmaOne^2)\gradtsq{} + \sigmaZero^2}} + \parens{2\sigmaZero \eta + \frac{L \eta^2}{2}}\Et{\frac{\sgradtsq{}}{b_{t-1}^2+ \sgradtsq{}}}
\end{align}
Summing this expression over all ``good'' times $t\in\Sg$, recalling that $\{t\in\Sg\}\in\F_{t-1}$, and applying the tower rule of expectations, we find that the LHS of the resulting expression can be written more simply as:
\begin{align*}
    \sum_{t\in[T]} \EXP{\Et{\ftplus - \ft} \1{t\in\Sg}}
    &=\sum_{t\in[T]} \EXP{\Et{(\ftplus - \ft)\1{t\in\Sg}}}\\
    &=\sum_{t\in[T]} \EXP{(\ftplus - \ft)\1{t\in\Sg}}\\
    &=\EXP{\sum_{t\in\Sg} \ftplus - \ft}.
\end{align*}
Thus, applying the same argument tower rule argument as above to the RHS of \eqref{eq:good_event_starting_pt} after summing over all $t\in\Sg$, and rearranging, we obtain
\begin{align*}
    \EXP{\sum_{t\in \Sg} \frac{\tetat}{4} \gradtsq{}}
    &\leq \EXP{\sum_{t\in\Sg} \ftplus - \ft}
        + \frac{\eta^2(\nicefrac{4\sigmaZero}{\eta} + L)}{2} \EXP{\sum_{t\in\Sg} \frac{\sgradtsq}{b_{t-1}^2 + \sgradtsq}}.
\end{align*}
Observing that, by adding and subtracting $\EXP{\sum_{t\not\in\Sg}\ft - \ftplus}$ to the above expression, and by upper bounding $\fzero - \EXP{\f{T}}\leq\fzero - \fstar$, we obtain the first inequality.

To obtain the second inequality, we note that, since $\braces{t\not\in\Sg}\in\F_{t-1}$, we may use the same arguments as presented earlier, along with the observation that, since $\nicefrac{\sgradtsq}{(b_{t-1}^2 + \sgradtsq)} \leq 1$ \emph{deterministically},
    $1-4\sigmaOne\sqrt{\Et{\nicefrac{\sgradtsq{}}{(b_{t-1}^2 + \sgradtsq{})}}}
    \geq 1-4\sigmaOne,$
to conclude that, whenever $t\not\in\Sg$,
\begin{align*}
    &\Et{\ftplus{} - \ft{}}\\
    &\leq
    \frac{\eta}{2}\parens{4\sigmaOne-1} \frac{\gradtsq{}}{\sqrt{b_{t-1}^2 + (1+\sigmaOne^2)\gradtsq{} + \sigmaZero^2}} + \parens{2\sigmaZero \eta + \frac{L \eta^2}{2}}\Et{\frac{\sgradtsq{}}{b_{t-1}^2+ \sgradtsq{}}}.
\end{align*}
Summing and taking expectations of the above expression, using the resulting expression to bound $\EXP{\sum_{t\not\in\Sg} \ftplus - \ft}$, and using \cref{cor:logBound} to bound $\EXP{\sum_{t=1}^T \frac{\sgradtsq}{b_{t-1}^2 + \sgradtsq}}\leq \log(f(T))$, we reach the desired inequality.
\end{custproof}

\begin{restatable}{custlemma}{restateGoodSet}\label{lem:goodSet}
    Let $\Sg$ be the set of ``good'' times from \cref{def:goodTimes}.
    Then, we have that{\color{cblue}, when $\sigmaOne \leq \nicefrac{1}{8}$, then $|\Sg^c| = 0$, and otherwise}\footnote{As an aside, using essentially the same arguments, we can show that $|\Sg^c|$ satisfies the Bernstein condition with parameter $\const \cdot \log(T)$, which implies that, with high probability, $|\Sg^c| \leq \const\cdot \log^2(T)$.}
    \begin{align*}
        \EXP{|\Sg^c|} \leq 64\sigmaOne^2\log(f(T))
        \quad \text{and} \quad
        \EXP{|\Sg^c|^2} \leq \parens{64\sigmaOne^2(1 + 128 \sigmaOne^2) + 2}\log^2(T^2 f(T)),
    \end{align*}
    where $f(\cdot)$ is as defined in \eqref{eq:polyFn}.
\end{restatable}

\begin{custproof}
{\color{cblue}
Observe that an equivalent condition for a time $t$ to be ``good'' in the sense of \cref{def:goodTimes} is:
\begin{align*}
    \Et{\frac{\sgradtsq}{b_{t-1}^2 + \sgradtsq}} \leq \frac{1}{64 \sigmaOne^2}.
\end{align*}
Whenever $\sigmaOne \leq \nicefrac{1}{8}$, the above inequality is (trivially) true deterministically since $\nicefrac{\sgradtsq}{ b_{t-1}^2 + \sgradtsq}\leq~1$, implying that $\Sg^c = \emptyset$. Thus, we will focus on the case when $\sigmaOne > \nicefrac{1}{8}$.
}

We first prove the first inequality.
Note that, if $t\not\in\Sg$, then $\Et{\nicefrac{\sgradtsq}{b_t^2}} > \nicefrac{1}{64\sigmaOne^2}$ by construction. Conveniently, this lower bound tells us that, for each time $t\in[T],$
\begin{align*}
    \EXP{\frac{\sgradtsq}{b_t^2}}
    = \EXP{\Et{\frac{\sgradtsq}{b_t^2}}}
    \geq \EXP{\Et{\frac{\sgradtsq}{b_t^2}}\1{t\not\in\Sg}}
    \geq \frac{\EXP{\1{t\not\in\Sg}}}{64\sigmaOne^2}.
\end{align*}
Now, summing the above expression over all times $t\in[T]$, and applying \cref{cor:logBound}, we find that
\begin{align*}
    \log(f(T)) 
    \geq \sum_{t\in[T]}\EXP{\frac{\sgradtsq}{b_t^2}}
    \geq \frac{1}{64\sigmaOne^2}\EXP{\sum_{t\in[T]}\1{t\not\in\Sg}}
    \geq \frac{1}{64\sigmaOne^2}\EXP{|\Sg^c|},
\end{align*}
as claimed.

Now, observe that, for that first result, we \emph{only} used our guarantee on $\EXP{\sum_{t\in[T]}\nicefrac{\sgradtsq}{b_0^2}}$. However, \cref{cor:logBound} tells us much more. Indeed, assuming that $\E_s(\delta)$ (the \emph{nice event} from \cref{def:niceEvents}) is true for some $s\in[T]$, {\color{cmaroon}and choosing (with foresight) $\delta = \nicefrac{1}{T^2}$,}
\begin{align}\label{eq:conditionalSetSize}
    \sum_{t=s+1}^T\Econd{s}{\1{t\not\in\Sg}}
    \leq 64\sigmaOne^2\log(\nicefrac{f(T)}{\delta}),
\end{align}
where the above follows by noting (similarly as before), for every $t > s$, {\color{cmaroon} since $\{t\not\in\Sg\}\in\F_{t-1}$, by an application of the tower rule of expectation and \cref{def:goodTimes}}
\begin{align}\label{eq:conditionalPigeonholing}
    \Econd{s}{\frac{\sgradtsq}{b_t^2}}
    &= \Econd{s}{\Et{\frac{\sgradtsq}{b_t^2}}}\nonumber\\
    &\geq \Econd{s}{\Et{\frac{\sgradtsq}{b_t^2}}\1{t\not\in\Sg}}
    \geq \frac{\Econd{s}{\1{t\not\in\Sg}}}{64\sigmaOne^2}.
\end{align}
{\color{cmaroon}Now, by \eqref{eq:logBoundHighProbCond} in \cref{cor:logBound}, we know that, whenever $\E_s(\delta)$ is true, then
\begin{align*}
    \Econd{s}{\sum_{t=s+1}^T \frac{\sgradtsq}{b_t^2}}
    \leq \Econd{s}{\sum_{t=1}^T \frac{\sgradtsq}{b_t^2}}
    \leq \log(\nicefrac{f(T)}{\delta}).
\end{align*}
Therefore, by summing \eqref{eq:conditionalPigeonholing} from $s+1$ to $T$ and rearranging, we obtain \eqref{eq:conditionalSetSize}.
}
We can use this bound as follows: since $|\Sg^c|^2 = \parens{\sum_{t\in[T]} \1{t\not\in\Sg}}^2$, we may expand this expression and apply the tower rule of expectations to observe that
\begin{align*}
    \EXP{|\Sg^c|^2}
    = \EXP{|\Sg^c|}
    +2 \sum_{t_1=1}^T \EXP{\1{t_1\not\in\Sg}\Econd{t_1}{\sum_{t_2=t_1+1}^T \1{t_2\not\in\Sg}}}.
\end{align*}
By \eqref{eq:conditionalSetSize}, we additionally know that, for each time $t_1 \leq T$,
\begin{align*}
    &\EXP{\1{t_1\not\in\Sg}\Econd{t_1}{\sum_{t_2=t_1+1}^T \1{t_2\not\in\Sg}}}\\
    &{\color{cmaroon}=\EXP{\1{t_1\not\in\Sg}\Econd{t_1}{\sum_{t_2=t_1+1}^T \1{t_2\not\in\Sg}}(\1{\E_{t_1}(\delta)} + \1{\E_{t_1}(\delta)^c})}}\\
    &\leq 64\sigmaOne^2\log(\nicefrac{f(T)}{\delta})\EXP{\1{t_1\not\in\Sg}}
    +T\PRO{\E_{t_1}(\delta)^c}. 
\end{align*}
As a result, since $\PRO{\E_{t_1}(\delta)^c}\leq \delta$ by construction, and {\color{cmaroon}by our choice of} $\delta=\nicefrac{1}{T^2}$, we conclude that
\begin{align*}
    \EXP{|\Sg^c|^2}
    &\leq (1 + 128\sigmaOne^2 \log(T^2 f(T)))\EXP{|\Sg^c|} + 2\\
    &\leq (64\sigmaOne^2 (1 + 128\sigmaOne^2 ) + 2)\log^2(T^2 f(T)),
\end{align*}
as claimed.
\end{custproof} \section{Compensating for ``Bad'' Time-Steps}
\label{sec:appendix:compensation}

Here, we provide proofs for the compensation arguments presented in \cref{sec:convergenceAdaGradNorm}

\begin{custlemma}\label{lem:summedBoundModifiedGoodSet}
In the same setting as \cref{lem:summedBoundGoodSet}, for any set $\Sc := \cup_{t\in\Sg^c} \Scomp{t} \subseteq \Sg$ (where $\Scomp{t}$ denotes the compensating set for a bad time $t$ which is disjoint from all other $\Scomp{\tilde{t}}$), we have that
\begin{align*}
    \EXP{\sum_{t\in \tSg} \frac{\tetat}{4}\gradtsq{}}
    &\leq \fzero - \fstar
    + \cCommon\log(f(T))\\
    &\quad+ \EXP{\sum_{t\not\in \Sg} \parens{\frac{(4\sigmaOne - 1)}{2} \tetat \gradtsq
    - \sum_{t'\in \Scomp{t}} \frac{\tetaat{t'}}{4} \gradsq{t'}}},
\end{align*}
where $\tSg := \Sg \setminus \Sc$ are the remaining ``good'' times after compensation, and $\cCommon=2\sigmaZero\eta + \nicefrac{L \eta^2}{2}$.
\end{custlemma}
\begin{custproof}
By subtracting 
$
    \EXP{\sum_{t'\in \Sc} \frac{\tetaat{t'}}{4}\gradsq{t'}}
$
from both sides of the expression in \cref{lem:summedBoundGoodSet} (since $\Sc \subseteq \Sg$) and using the fact that the $\Scomp{t}$ partition $\Sc$, the claimed inequality is immediate.
\end{custproof}

\restateCompensationInsight*

\begin{custremark}[On the interpretation of and proof techniques for \cref{lem:compensationInsight}]
\label{rem:compensationInsight}
Note that we will use \cref{lem:compensationInsight} in order to bound (some of) the residual terms in \cref{lem:summedBoundModifiedGoodSet}, and thus, in that context, will take $t$ to be some ``bad'' time, and $\Scomp{t}$ to be the set of $\ncomp$ compensating ``good'' times for $t$.
We emphasize, however, that the proof of \cref{lem:compensationInsight} does not rely on the notions of ``good'' or ``bad'' times from \cref{def:goodTimes}. Indeed, this result holds true for \emph{any} time $t\in[T]$ and set $\Scomp{t}$ which satisfies conditions (i) and (ii) from the statement. The proof will exploit special properties of the algorithm \eqref{eq:alg} and the smoothness of the objective function, and holds deterministically.
\end{custremark}

\begin{custproof}
Let us begin by proving that, for any times $t \geq t'$,
\begin{align}\label{eq:compInsightStep1}
    \frac{\tetaat{t}}{4}\gradsq{t} - \tetaat{t'}\gradsq{t'} \leq \frac{\eta^2 L (t - t')}{2}.
\end{align}
The claim is trivial when $t'=t,$ so we focus on the case when $t' < t$.
Let us denote $a=b_{t-1}^2 + (1+\sigmaOne^2)\gradsq{t}+\sigmaZero^2$ and $b=b_{t'-1}^2 + (1+\sigmaOne^2)\gradsq{t'}+\sigmaZero^2$. Then, observe that
\begin{align*}
    \frac{1}{\sqrt{a}} - \frac{1}{\sqrt{b}}
    =
    \frac{\sqrt{b}-\sqrt{a}}{\sqrt{a b}}
    =
    \frac{b-a}{\sqrt{a b}(\sqrt{a} + \sqrt{b})}.
\end{align*}
Therefore, we can observe that the step sizes are sufficiently close, since
\begin{align*}
    \frac{\tetaat{t} - \tetaat{t'}}{\eta} 
    &= \frac{1}{\sqrt{b_{t-1}^2 + (1+\sigmaOne^2)\gradsq{t} + \sigmaZero^2}} - \frac{1}{\sqrt{b_{t'-1}^2 + (1+\sigmaOne^2)\gradsq{t'} + \sigmaZero^2}}\\
    &\leq \frac{(1+\sigmaOne^2)\parens{\gradsq{t'} - \gradsq{t}}}{\sqrt{ab}\parens{\sqrt{a} + \sqrt{b}}}\\
    &= \frac{(1+\sigmaOne^2)\parens{\gradnorm{t'} - \gradnorm{t}}\parens{\gradnorm{t'} + \gradnorm{t}}}{\sqrt{ab}\parens{\sqrt{a} + \sqrt{b}}}\\
    &\leq \frac{(1+\sigmaOne^2)\abs{\gradnorm{t'} - \gradnorm{t}}}{\sqrt{ab}}\\
    &\leq \frac{\eta L(t-t')}{\gradnorm{t}\gradnorm{t'}}
\end{align*}
where the last line follows by \cref{lem:trivialGradBound}. We will now use this observation in order to prove the claimed inequality. We will proceed by considering two cases.

In the first case, if $\gradnorm{t} > 2\eta L (t-t')$, then by \cref{lem:trivialGradBound}, $\gradnorm{t'} \geq \gradnorm{t} - \eta L (t-t')\geq \nicefrac{1}{2}\gradnorm{t}$. This implies that
\begin{align*}
   \frac{1}{4}\tetaat{t}\gradsq{t} - \tetaat{t'}\gradsq{t'}
&\leq
   \frac{1}{4}\gradsq{t}\parens{\tetaat{t} - \tetaat{t'}}\\
   &\leq
   \frac{\eta^2 L (t-t')\gradnorm{t}}{4\gradnorm{t'}}\\
   &\leq
   \frac{\eta^2 L (t-t')}{2}
\end{align*}
In the alternative case, when $\gradnorm{t} \leq 2\eta L (t-t')$, then
\begin{align*}
   \frac{1}{4}\tetaat{t}\gradsq{t} - \tetaat{t'}\gradsq{t'}
   \leq
   \frac{1}{4}\tetaat{t}\gradsq{t}
   \leq
   \frac{\eta}{4}\gradnorm{t}
   \leq
   \frac{\eta^2 L (t-t')}{2}
\end{align*}
where the first inequality follows by lower bounding the second term by zero, and the second by definition of $\tetaat{t}$, and the third by assumption.
Thus, we obtain exactly the same bound in both cases, which establishes \eqref{eq:compInsightStep1}.
Now, we can use \eqref{eq:compInsightStep1} to prove the claim. Indeed, since, by construction, $|\Scomp{t}| = \ncomp \geq 8(4\sigmaOne-1)$ we have
\begin{align*}
    \frac{4\sigmaOne - 1}{2}\tetat\gradtsq - \sum_{t'\in \Scomp{t}}\frac{\tetaat{t'}}{4}\gradsq{t'}
    &\leq \frac{1}{4}\sum_{t'\in \Scomp{t}} \parens{\frac{\tetat}{4}\gradtsq - \tetaat{t'}\gradsq{t'}}.
\end{align*}
Therefore, using \eqref{eq:compInsightStep1} to bound the above, and recalling that $|S|=\ncomp$, we conclude that
\begin{align*}
    \frac{4\sigmaOne - 1}{2}\tetat\gradtsq - \sum_{t'\in \Scomp{t}}\frac{\tetaat{t'}}{4}\gradsq{t'}
    \leq \frac{\eta^2 L \ncomp}{8}(t-\min(\Scomp{t})),
\end{align*}
as claimed.
\end{custproof}
\restateCompSetExists*

\begin{custproof}
~\paragraph{\underline{Constructing $\Sc$.}}
We begin by giving a detailed description of our \emph{greedy} construction of $\Sc$ which was briefly described in \cref{sec:convergenceAdaGradNorm}. To begin, let us denote $\tb{i}$ as the $i$th largest time in $\Sg^c$. For notational simplicity, we will abuse our notation and refer to $\Scomp{i}$ and $\Scomp{\tb{i}}$ interchangably as the set of compensating ``good'' times for $\tb{i}$.  We will iteratively construct each $\Scomp{i}$ for each $i \in [|\Sg^c|]$, starting with $i=1$. Let us denote
\begin{align*}
    \Sel{i} = \{t \in \Sg \mid t < \min\{\tb{i}, \min(\Scomp{i-1})\}\}
\end{align*}
as the set of \emph{eligible} compensating ``good'' times for $\tb{i}$. Intuitively, these are the set of ``good'' times \emph{smaller} than $\tb{i}$ which have not been used to compensate for \emph{larger} bad times $\tb{i'} > \tb{i}$. Note that we take $\min(\Scomp{0}) = +\infty$ and $\min(\emptyset) = - \infty$ so that (i) $\Sel{i}$ consists of \emph{every} ``good'' time which is smaller than $\tb{i}$, and (ii) if $\Scomp{i-1} = \emptyset$, then there are \emph{no} eligible times for $\tb{i}$, i.e., $\Sel{i}=\emptyset$.

We may then choose the ``compensating'' set $\Scomp{i}$ for $\tb{i}$ as the \emph{largest} (at most) $\ncomp$ times in $\Sel{i}$. It is clear by this construction that $\Scomp{i} \cap \Scomp{i'} = \emptyset$ for every $i \neq i' \in [|\Sg^c|]$.

We will further take $i^*$ to be the \emph{smallest} index in $[|\Sg^c|]$ such that $|\Scomp{i}| < \ncomp$. Intuitively, this is the index of the \emph{largest} ``bad'' time $\tb{i}$ which is not \emph{fully} compensated.

\paragraph{\underline{Establishing the properties of $\Sc$.}}
Note that, as required, each $|\Scomp{i}| \leq \ncomp$ and $\tb{i} > \max(\Scomp{i})$ by the construction of $\Sc$ described above. {\color{cblue} Additionally, whenever $\ncomp = 0$, the result is immediately true, so we proceed assuming that $\ncomp > 0$.} Further, note that since $i^*$ is chosen as the \emph{smallest} index for which $|\Scomp{i^*}| < \ncomp$, it must be the case that $|\Scomp{i}| = \ncomp$ for every $i < i^*$, and $|\Scomp{i}| < \ncomp$ for every $i\geq i^*$. Therefore, to reason about the two conditions, we need to consider only the cases (i) $\tb{i} > \tb{i^*}$ and (ii) $\tb{i} \leq \tb{i^*}$.

\noindent\textit{Case 1:}
Let us first consider a bad time $\tb{i} > \tb{i^*}$. Clearly, $|\Scomp{i}|=\ncomp$.
By the greedy construction of the compensating sets, observe that
\begin{align}\label{eq:compConstruction1}
    \abs{\left(\max(\Scomp{i}), \tb{i}\right) \cap \Sg} \leq (i-1) \cdot \ncomp.
\end{align}
Indeed, these are the times in $\Sg$ associated with a compensating set $\Scomp{i'}$ for a \emph{larger} ``bad'' time $\tb{i'} > \tb{i}$. If there were any more ``good'' times on this interval, then they would have been assigned to $\Scomp{i}$ by definition of our greedy procedure. Next, note that
\begin{align}\label{eq:compConstruction2}
    \abs{\left[\min(\Scomp{i}), \max(\Scomp{i})\right] \cap \Sg} = \ncomp.
\end{align}
These times corresponding to the $\ncomp$ times in $\Scomp{i}$. Indeed, by the greedy construction of our compensating sets, $\max(\Scomp{i'+1}) < \min(\Scomp{i'})$ for every $i'\in[|\Sg^c|]$, and the procedure always chooses the largest ``good'' times available in $\Sel{i}$, so no other good times can lie on this interval. Finally, we observe that
\begin{align}\label{eq:compConstruction3}
    \abs{\left[\min(\Scomp{i}), \tb{i}\right) \cap \Sg^c} \leq |\Sg^c| - i,
\end{align}
corresponding to the \emph{at most} $|\Sg^c| - i$ bad times $\tb{i'} < \tb{i}.$ Combining \cref{eq:compConstruction1,eq:compConstruction2,eq:compConstruction3}, we conclude that $\tb{i} - \min(\Scomp{i}) \leq \ncomp \cdot |\Sg^c|$.

\noindent\textit{Case 2:}
We now consider the case when $\tb{i} \leq \tb{i^*}$. Clearly, $|\Scomp{i}| < \ncomp$.
Since we need only to show that $\tb{i}$ is upper bounded by $\ncomp\cdot|\Sg^c|$, it suffices to show this for $\tb{i^*}$.
Our arguments will follow in a similar spirit as \textit{Case 1}. Indeed, using exactly the same arguments used to establish \cref{eq:compConstruction1,eq:compConstruction2}, we know that
\begin{align}\label{eq:compConstruction4}
    \abs{\left[\min(\Scomp{i^*}), \tb{i}\right) \cap \Sg} \leq i^* \cdot \ncomp.
\end{align}
Further, by the greedy construction of the compensating sets, since $|\Scomp{i^*}| < \ncomp$, it \emph{must} be the case that
\begin{align}\label{eq:compConstruction5}
    \left[1, \min(\Scomp{i^*})\right) \cap \Sg = \emptyset,
\end{align}
since otherwise, any remaining elements could have been added to $\Scomp{i^*}$. Therefore, since
\begin{align}\label{eq:compConstruction6}
    \abs{\left[1,\tb{i^*}\right) \cap \Sg^c} = |\Sg^c| - i^*,
\end{align}
we conclude by \cref{eq:compConstruction4,eq:compConstruction5,eq:compConstruction6} that $\tb{i^*} \leq \ncomp \cdot |\Sg^c|$, as claimed.
\end{custproof}

\begin{restatable}{custlemma}{restateResidualBound}\label{lem:residualBound}
If $\Sc$ is constructed as in \cref{lem:compSetExists}, then the ``residual'' term from \cref{lem:summedBoundModifiedGoodSet} can be bounded as follows:
\begin{align*}
    &\EXP{\sum_{t\not\in \Sg} \parens{\frac{(4\sigmaOne - 1)}{2} \tetat \gradtsq
    - \sum_{t'\in \Scomp{t}} \frac{\tetaat{t'}}{4} \gradsq{t'}}}\\
    &\leq
    128\eta\sigmaOne^2 \gradzeronorm \log(f(T)){\color{cblue}\1{\sigmaOne > \nicefrac{1}{8}}}\\
    &\quad+\eta^2 L \ncomp\parens{\frac{\ncomp}{8}+2}(64 \sigmaOne^2(1+128\sigmaOne^2) + 2) \log^2(T^2 f(T)),
\end{align*}
{\color{cblue}where $\ncomp = \max\{8\ceil{4\sigmaOne - 1},0\}$ is as in \cref{lem:compSetExists}.}
\end{restatable}

\begin{custproof}
Borrowing the notation from the proof of \cref{lem:compSetExists}, we will use $\tb{i}$ to denote the $i$th largest ``bad'' time in $\Sg$, and, abusing notation slightly, use $\Scomp{i}$ and $\Scomp{\tb{i}}$ interchangeably to denote the compensating ``good'' times for $\tb{i}$. {\color{cblue}For the purpose of this proof, we may assume that $\sigmaOne > \nicefrac{1}{4}$ (which also implies $\ncomp > 0$), since otherwise the result is trivially true because the left-hand side of the claimed inequality is negative in this case.} Further, we take $i^*$ to be the index of the first ``bad'' time $\tb{i^*}$ which cannot be fully compensated, i.e., $|\Scomp{i^*}| <  \ncomp$.
Using this notation, we may rewrite the residual term from \cref{lem:summedBoundModifiedGoodSet} in the following convenient manner:
\begin{align*}
    &\EXP{\sum_{t\not\in \Sg} \parens{\frac{(4\sigmaOne - 1)}{2} \tetat \gradtsq
    - \sum_{t'\in \Scomp{i}} \frac{\tetat}{4} \gradtsq}}\\
    &=
    \EXP{\sum_{i < i^*} \parens{\frac{(4\sigmaOne - 1)}{2} \tetaat{\tb{i}} \gradsq{\tb{i}}
    - \sum_{t'\in \Scomp{i}} \frac{\tetaat{t'}}{4} \gradtsq}}\\
    &\quad + 
    \EXP{\sum_{i \geq i^*} \frac{(4\sigmaOne - 1)}{2} \tetaat{\tb{i}} \gradsq{\tb{i}}}.
\end{align*}
Now, we will use \cref{lem:compensationInsight} to bound the first term above. We will use the trivial bound for the second term: by \cref{def:stepSizeProxy} and \cref{lem:trivialGradBound}, we may bound each term inside of the sum of the second expression above as:
\begin{align*}
    \tetaat{\tb{i}} \gradsq{\tb{i}} 
    \leq \frac{\eta}{\sqrt{1+\sigmaOne^2}} \gradnorm{\tb{i}} 
    \leq \frac{\eta}{\sqrt{1+\sigmaOne^2}} (\gradzeronorm + \eta L \tb{i})
\end{align*}
These two bounds described above, together with \cref{lem:compSetExists} and the fact that each $i\leq |\Sg^c|$, imply that:
\begin{align*}
    &\EXP{\sum_{t\not\in \Sg} \parens{\frac{(4\sigmaOne - 1)}{2} \tetat \gradtsq
    - \sum_{t'\in \Sc} \frac{\tetaat{t'}}{4} \gradsq{t'}}}\\
    &\leq \EXP{\sum_{i < i^*} \frac{\eta^2 L \ncomp}{8}(\tb{i} - \min(\Scomp{i}))}
    + \EXP{\sum_{i \geq i^*} \eta\frac{4\sigmaOne-1}{2\sqrt{1+\sigmaOne^2}}(\gradzeronorm + \eta L \tb{i})}\\
    &\leq \frac{\eta^2 L \ncomp^2}{8}\EXP{|\Sg^c|^2}
    + 2\eta \parens{\gradzeronorm\EXP{|\Sg^c|} + \eta L \ncomp\EXP{|\Sg^c|^2}}\\
    &\leq \eta^2 L \ncomp \parens{\frac{\ncomp}{8} + 2}\EXP{|\Sg^c|^2}
    + 2\eta\gradzeronorm \EXP{|\Sg^c|}.
\end{align*}
Applying the bounds on $|\Sg^c|$ from \cref{lem:goodSet} yields the claimed bound.
\end{custproof}

\begin{custlemma}\label{lem:mainSummedBoundAffineInterpreted}
Let the set $\Sc$ from \cref{lem:summedBoundModifiedGoodSet} be chosen as in \cref{lem:compSetExists}. Then, taking $\tSg := \Sg\setminus \Sc$ as the remaining ``good'' times after compensation, we have that
{\color{cblue}
\begin{align*}
    \EXP{\sum_{t\in \tSg} \frac{\tetat}{4} \gradtsq{}}
    \leq \fzero - \fstar
    + \tcCommon\cdot\log(f(T))
    + \tconstInvariance \cdot \log^2(T^2 f(T)),
\end{align*}
where we can take
\begin{align*}
    \tcCommon 
    = \cCommon + 128\eta\sigmaOne^2\gradzeronorm\1{\sigmaOne > \nicefrac{1}{8}}
    \quad \text{where}\quad
    \cCommon = 2\eta\sigmaZero + \frac{L \eta^2}{2},
\end{align*}
and
\begin{align*}
    \tconstInvariance = L\eta^2 \ncomp (\nicefrac{\ncomp}{8}+2)(64\sigmaOne^2 +8192\sigmaOne^4 + 2)
    \quad \text{where}\quad
    \ncomp = \max\{8\ceil{4\sigmaOne - 1},0\}.
\end{align*}
In particular, we have that
\begin{align*}
    \EXP{\sum_{t\in \tSg} \frac{\tetat}{4} \gradtsq{}}
    \leq \fzero - \fstar
    + \constInvariance \cdot \log^2(T^2 f(T)),
\end{align*}
where $\constInvariance = \tcCommon + \tconstInvariance.$
}
\end{custlemma}
\begin{custproof}
The result follows immediately by combining \cref{lem:residualBound,lem:summedBoundModifiedGoodSet}. Note that this result, up to logarithmic factors, takes essentially the same form as in the uniformly-bounded setting \eqref{eq:standardAnalysisRewrittenExpSummed}.
\end{custproof}

 \section{Bounding the Expected Sum of Gradients via Recursive Improvement}
\label{sec:appendix:recursiveInterlacing}

Here, we provide a proof for the recursive improvement argument presented in \cref{sec:convergenceAdaGradNorm}.

\subsection{Main Ideas}
{
\begin{custlemma}\label{lem:recursiveInterlacing}
Suppose that{\color{cblue}, for some constants $x\geq 1$ and $y\geq 0$, the following inequality is true:}
\begin{align}\label{eq:polyGradSummedBoundAssumed}
    \EXP{\sum_{t\in[T]} \gradtsq} \leq \constRecImp T^x \log^y(T^2 f(T)),
\end{align}
{where} $\constRecImp$ is specified as
{\color{cblue}
\begin{align*}
    \constRecImp = \max\bigg\{
    b_0^2 + \sigmaZero^2 + 32(1+8(1+\ncomp)\sigmaOne^2)(\gradzerosq + \eta^2 L^2),
    512\parens{\frac{\fzero - \fstar + \constInvariance}{\eta}}^2
    \bigg\},
\end{align*}
with $\ncomp = \max\{8\ceil{4\sigmaOne - 1},0\}$ defined in \cref{lem:compensationInsight} and $\constInvariance$ the constant defined in \cref{lem:mainSummedBoundAffineInterpreted},
\begin{align*}
    \constInvariance = 2\eta\sigmaZero + \frac{L\eta^2}{2}
    + 128\eta\sigmaOne^2\gradzeronorm\1{\sigmaOne > \nicefrac{1}{8}}
    + L \eta^2 \ncomp(\nicefrac{\ncomp}{8} + 2)(64\sigmaOne^2 + 8192\sigmaOne^4 + 2).
\end{align*}
}

Then, in fact, the following tighter bound also holds:
\begin{align}\label{eq:interlacingResult1}
    \EXP{\sum_{t\in[T]} \gradtsq} 
    \leq 
    \constRecImp T^{{\color{cblue}\max\braces{\frac{x + 2}{3},\frac{x}{2}}}}\log^{{\color{cblue}\max\braces{\frac{y + 5}{3},2}}}(T^2 f(T)).
\end{align}
In particular, as a consequence of \cref{lem:logGradBound,cor:logBound},
\begin{align}\label{eq:interlacingResult2}
    \EXP{\sum_{t\in[T]} \gradtsq} \leq \constRecImp T \log^{\frac{5}{2}}(T^2 f(T)).
\end{align}
\end{custlemma}

The main idea of the proof is to recursively improve our upper bound on the ``normalized'' expected sum of gradients from \cref{lem:mainSummedBoundAffineInterpreted} \emph{in expectation} combining it with a lower bound on the step size proxy \emph{with high (enough) probability} obtained from Markov's inequality and an invariant upper bound provided in \cref{lem:informalMainSummedBoundAffineInterpreted}. Recall that  $\tetaat{T} = \nicefrac{\eta}{\sqrt{b_{T-1}^2 + (1+\sigmaOne^2)\gradtsq + \sigmaZero^2}}$, thus to provide a lower bound on the step size proxy we will focus on \emph{upper} bounding $b_{T-1}$. In particular taking the expectation, we have that:
\begin{align}\label{eq:recursiveInterlacing:btBound}
    \EXP{b_{T-1}^2} 
    = b_0^2 + \sum_{t=1}^{T-1}\EXP{\sgradtsq}
    \leq b_0^2 + (T-1)\sigmaZero^2 + (1+\sigmaOne^2) \sum_{t=1}^{T-1}\EXP{\gradtsq},
\end{align}
where the above follows by applying \cref{assump:unbiasedGrad,assump:affineVariance}. Thus, to obtain an upper bound for $\EXP{b_{T-1}^2}$, we must have a bound for $\EXP{\sum_{t\in[T-1]}\gradtsq}$ -- the quantity we wish to bound! This highlights the motivation for applying the following improving idea \emph{recursively}. We begin with a crude (polynomial in $T$) upper bound for $\EXP{\sum_{t\in[T]}\gradtsq}$, and recursively \emph{improve} this bound via the interlaced inequalities described above. 
Repeating this process infinitely many times ultimately obtains the desired upper bound on the expected sum of the gradients.

\begin{custproof}[of \cref{lem:recursiveInterlacing}]
The proof will proceed in three steps, in which we will invoke the auxiliary \cref{lem:interlacingStepSizeLB,lem:removingTheIndicator,lem:interlacingBadTerms}. It is straightforward to verify that the constant $\constRecImp$ specified in this lemma, as well as the choice of $h(T) = T^2 f(T)$, satisfy the constraints from those lemmas. Thus, we are free to use these results to prove our desired result.
~\paragraph{Step 1:  Lower bounding the step size proxy.}
 Recall that \cref{lem:mainSummedBoundAffineInterpreted} gives an upper bound on $\EXP{\sum_{t\in\Sg\setminus\Sc}\tetat\gradtsq}$. 
Using the ``nice event'' $\E_T(\delta)$ from \cref{def:niceEvents} with a \emph{sufficiently small failure probability} $\delta = \nicefrac{({\color{cblue}2} + \sigmaOne^2)\log^{y-\gamma_2}(T^2 f(T))}{T^{\gamma_1}}$, where $\gamma_1,\gamma_2 \geq 0$ are arbitrary parameters satisfying $\gamma_1 + x \geq 2$ and $\gamma_1 \leq {\color{cblue}1},$ {\color{cblue} $y\geq\gamma_2$, and $x$ and $y$ are the parameters from \eqref{eq:polyGradSummedBoundAssumed}.} 
we can ensure that the step size proxy $\tetat$ is \emph{sufficiently small}. Indeed, these insights allow us to prove \cref{lem:interlacingStepSizeLB}, which tells us that:
\begin{align}\label{eq:interlacing1}
    \EXP{\sum_{t\in\tSg} \tetat \gradtsq}
    \geq \EXP{\sum_{t\in\tSg} \tetat \gradtsq \1{\E_T(\delta)}}
    \geq \frac{\eta\EXP{\sum_{t\in\tSg} \gradtsq \1{\E_T(\delta)}}}{\sqrt{2 \constRecImp T^{x+\gamma_1}\log^{\gamma_2}(T^2 f(T))}}.
\end{align}
While the above translates the bound in \cref{lem:mainSummedBoundAffineInterpreted} into a more interpretable form, the presence of $\1{\E_T(\delta)}$ makes the above bound not immediately useful. However, by construction, $\E_T(\delta)$ happens with probability at least $1-\delta$. Our choice of $\delta$ will allow us to show that, effectively, the above upper bound is still true with the indicator removed.

In order to ``remove'' the indicator from the expectation above, we will need to show that, when $\E_T(\delta)$ is false, $\sum_{t\in\tSg} \gradtsq$ cannot be \emph{too large}.
Recall that we have two main tools to upper bound the size of this sum: \cref{lem:trivialGradBound}, which gives a \emph{deterministic} upper bound of $\O(T^3)$, and \cref{lem:logGradBound}, which gives a \emph{high-probability} upper bound of $\tO(T^2)$. These insights allow us to prove \cref{lem:removingTheIndicator}, which tells us that
\begin{align}\label{eq:interlacing2}
    \EXP{\sum_{t\in\tSg} \gradtsq \1{\E_T(\delta)}}
    \geq \EXP{\sum_{t\in\tSg} \gradtsq}
    - \frac{\constRecImp}{4} T^{2-\gamma_1}\log^{y-\gamma_2+1}(T^2 f(T)).
\end{align}

\paragraph{Step 2: Bounding the ``good'' terms.}
With the indicator removed from the above expression, we are now ready to use \cref{lem:mainSummedBoundAffineInterpreted} together with \eqref{eq:interlacing1} and \eqref{eq:interlacing2} to obtain a bound on the expected size of the gradients \emph{at the good times}:
\begin{align*}
    \EXP{\sum_{t\in\tSg} \gradtsq}
    &\leq \parens{4\sqrt{2 \constRecImp T^{x+\gamma_1} \log^{\gamma_2}(T^2 f(T))}}\frac{\fzero - \fstar
    + \constInvariance \log^2(T^2 f(T))}{\eta}\\
    &\quad+ \frac{\constRecImp}{4} T^{2-\gamma_1}\log^{y-\gamma_2+1}(T^2 f(T))\\
    &\leq 
    \frac{\constRecImp}{4} T^{\frac{x + \gamma_1}{2}}\log^{2 + \frac{\gamma_2}{2}}(T^2 f(T))\\
    &\quad+ \frac{\constRecImp}{4} T^{2-\gamma_1}\log^{y-\gamma_2+1}(T^2 f(T)),
\end{align*}
where the second inequality follows by upper bounding 
{\color{cblue}$4 \sqrt{2}\nicefrac{(\fzero - \fstar + \constInvariance)}{\eta} \leq \nicefrac{\sqrt{\constRecImp}}{4}.$} Hence, by choosing $\gamma_1 = \max\{\nicefrac{(4-x)}{3},0\}$ and $\gamma_2 = \max\{\nicefrac{2(y-1)}{3},0\}$\footnote{Note that these choices of $\gamma_1,\gamma_2$ satisfy the requirements of \cref{lem:interlacingStepSizeLB,lem:removingTheIndicator,lem:interlacingBadTerms}. Indeed, $\gamma_1, \gamma_2\geq 0$ by construction. Further, since $x\geq 1$, we have that $\gamma_1 \leq \nicefrac{(4-x)}{3} \leq 1$ and, whenever $x\in [1,4]$, $x+\gamma_1 = \nicefrac{(2x+4)}{3} \geq 2$, and when $x>4$, $x+\gamma_1 = x > 4 > 2$. Finally, $y - \gamma_2 = \min\{\nicefrac{(y+2)}{3},y\}\geq 0$ since $y\geq 0$.}, we conclude that
\begin{align}\label{eq:interlacing3}
    \EXP{\sum_{t\in\tSg} \gradtsq}
    &\leq 
    \frac{\constRecImp}{2} T^{{\color{cblue}\max\braces{\frac{x + 2}{3},\frac{x}{2}}}}\log^{{\color{cblue}\max\braces{\frac{y + 5}{3},2}}}(T^2 f(T)).
\end{align}

\paragraph{Step 3: Bounding the ``bad'' terms.}
To conclude the argument, we will need to bound the remaining terms, $\EXP{\sum_{t\not\in\tSg} \gradtsq}$. Intuitively, these terms are not problematic for the sake of this argument, since (i) $\EXP{|\tSgc|} = \EXP{|\Sg^c \cup \Sc|} \leq (1+\ncomp) \EXP{|\Sg^c|} \leq (1+\ncomp)64\sigmaOne^2 \log(f(T))$ by construction of $\Sc$ (\cref{lem:compSetExists}) and by our control on the ``good'' set in \cref{lem:goodSet}, and since (ii) each term $\gradtsq$ can be bounded with high probability by $\O(T\log({\color{cmaroon}T^2}f(T)))$ by \cref{lem:logGradBound}. These arguments are formalized in \cref{lem:interlacingBadTerms}, which tells us that
\begin{align}\label{eq:interlacing4}
    \EXP{\sum_{t\not\in\tSg} \gradtsq}
    &\leq 
    \frac{\constRecImp}{2} T\log^{2}(T^2 f(T)).
\end{align}
Thus, we arrive at \eqref{eq:interlacingResult1} by combining the results of \eqref{eq:interlacing3} and \eqref{eq:interlacing4}, using the fact that $1 \leq \max\{\nicefrac{(x+2)}{3},\nicefrac{x}{2}\}$ since $x\geq 1$. 
To obtain \eqref{eq:interlacingResult2}, simply note that we may initialize \eqref{eq:interlacingResult1} with $x=2$ and $y=1$ by \cref{lem:logGradBound,cor:logBound}{\color{cblue}, since these Lemmas imply that $\EXP{\sum_{t\in[T]}\gradtsq} \leq 2(\gradzerosq + \eta^2 L^2)T^2 \log(f(T)) \leq \constRecImp T^2 \log(T^2 f(T))$. Alternatively, we could choose $x=3$ and $y=0$ by \cref{lem:trivialGradBound}, which implies that $\EXP{\sum_{t\in[T]}\gradtsq} \leq 2(\gradzerosq + \eta^2 L^2)T^3 \leq \constRecImp T^3$.} 
Given {\color{cblue}either of these initializations}, we may invoke our improved bound on the expected sum of gradients \eqref{eq:interlacingResult1} recursively, concluding that we may take $x=1$ and $y=\nicefrac{5}{2}$, as claimed.
\end{custproof}
}

\subsection{Technical Lemmas}

\begin{custlemma}[Polynomial control of step sizes]\label{lem:interlacingStepSizeLB}
Suppose that:
\begin{align}\label{eq:assumedPolyGradBound}
    \EXP{\sum_{t\in [T]}\gradtsq} \leq \constRecImp T^{x}\log^y(h(T))
\end{align}
for some $x,y\geq 1$, and 
{\color{cblue} $\constRecImp \geq \max\braces{b_0^2 + \sigmaZero^2, (1+\sigmaOne^2)(\gradzeronorm + \stepBndAGNorm L)^2 + \sigmaZero^2}$} and $h(T) \geq {\color{cblue}e}$.
Recalling the ``nice'' event $\E_T(\delta)$ from \cref{def:niceEvents}, where we choose 
{\color{cblue} $\delta = \frac{(2+\sigmaOne^2)\log^{y-\gamma_2}(h(T))}{T^{\gamma_1}}$}
for any $\gamma_1,\gamma_2$ satisfying $\gamma_1 + x\geq 2$, $\gamma_1,\gamma_2 \geq 0$, $\gamma_1 \leq 1$ $y\geq \gamma_2$. 
Then, recalling the step size proxy from \cref{def:stepSizeProxy}, $\tetat$, we have that
\begin{align}\label{eq:targetInterlacingStepSizeLB}
    \EXP{\sum_{t\in\tSg} \tetat\gradtsq}
    \geq
    \frac{\eta \EXP{\sum_{t\in\tSg} \gradtsq \1{\E_T(\delta)}}}{\sqrt{2 \constRecImp T^{x+\gamma_1}\log^{\gamma_2}(h(T))}}.
\end{align}
\end{custlemma}
\begin{custproof}
{\color{cblue}
We divide the proof in two cases: (1) $\delta > 1$, and (2) $\delta \leq 1$. In the first case, the claimed result \eqref{eq:targetInterlacingStepSizeLB} holds trivially, since $\E_T(\delta) = \emptyset$ by definition (see \cref{def:niceEvents}), and thus,
\begin{align*}
    \EXP{\sum_{t\in\tSg}\tetat\gradtsq}
    \geq 0
    = \frac{\eta \EXP{\sum_{t\in\tSg} \gradtsq \1{\E_T(\delta)}}}{\sqrt{2 \constRecImp T^{x+\gamma_1}\log^{\gamma_2}(h(T))}},
\end{align*}
since $\1{\E_T(\delta)} = \1{\emptyset} = 0$ deterministically. Thus, for the remainder of the proof, we assume that $\delta \leq 1$.
}

Let us assume that $\E_T(\delta)$ (the ``nice'' event from \cref{def:niceEvents}) is true. 
{\color{cblue}
Then, we have that
\begin{align*}
    b_T^2
    &\leq b_0^2 + \frac{T \sigmaZero^2 + (1+\sigmaOne^2) \constRecImp T^x \log^y(h(T))}{\delta}\\
    &\leq \frac{b_0^2 + T \sigmaZero^2 + (1+\sigmaOne^2) \constRecImp T^x \log^y(h(T))}{\delta}\\
    &\leq \frac{\parens{\frac{b_0^2 + \sigmaZero^2}{\constRecImp} + (1+\sigmaOne^2)} \constRecImp T^x \log^y(h(T))}{\delta}\\
    &\leq \constRecImp T^{x+\gamma_1} \log^{\gamma_2}(h(T)),
\end{align*}
where the first inequality follows by definition of $\E_T(\delta)$ and by the assumed bound \eqref{eq:assumedPolyGradBound}. The second inequality follows since $1 \leq \nicefrac{1}{\delta}$, and the third since $x\geq 1$ and $\log(h(T))\geq 1$. The final inequality follows by plugging in our choice of $\delta$, and using the fact that {\color{cblue}$\constRecImp \geq b_0^2 + \sigmaZero^2$.}
}

Now, since $b_{t-1}^2 \leq b_T^2$, the above inequality implies that
\begin{align*}
    \frac{\eta}{\tetat}
    &= \sqrt{b_{t-1}^2 + (1+\sigmaOne^2)\gradtsq + \sigmaZero^2}\\
    &\leq \sqrt{\constRecImp T^{x+\gamma_1}\log^{\gamma_2}(h(T)) + (1+\sigmaOne^2)(\gradzeronorm + \stepBndAGNorm L T)^2 + \sigmaZero^2}\\
    &\leq \sqrt{2 \constRecImp T^{x+\gamma_1}\log^{\gamma_2}(h(T))},
\end{align*}
where the first inequality follows since $b_{t-1}^2 \leq b_{T}^2$ and by \cref{lem:trivialGradBound}, and the second since $x+\gamma_1 \geq 2$ and {\color{cblue}$\constRecImp \geq (1+\sigmaOne^2)(\gradzeronorm + \stepBndAGNorm L)^2 + \sigmaZero^2$}. 

Noting that $\EXP{\sum_{t\in\tSg} \tetat\gradtsq} \geq \EXP{\sum_{t\in\tSg} \tetat\gradtsq \1{\E_T(\delta)}}$, and using the lower bound derived above, we obtain the claimed lower bound of \eqref{eq:targetInterlacingStepSizeLB}.
\end{custproof}
\begin{custlemma}[Removing $\1{\E_T(\delta)}$]\label{lem:removingTheIndicator}
Let us consider same setting as in \cref{lem:interlacingStepSizeLB}, assuming additionally that 
{\color{cblue} $\constRecImp \geq 16(2+\sigmaOne^2)(\gradzerosq + L^2\eta^2)$} and $h(T) \geq T^2 f(T)$. Then, recalling the set $\tSg = \Sg \setminus \Sc$ as constructed in \cref{lem:compSetExists}, we have that
\begin{align*}
    \EXP{\sum_{t\in\tSg} \gradtsq \1{\E_T(\delta)}}
    \geq~&\EXP{\sum_{t\in\tSg} \gradtsq}\\
    &- \frac{\constRecImp}{4} T^{2-\gamma_1}\log^{y-\gamma_2+1}(h(T))
\end{align*}
\end{custlemma}
\begin{custproof}
Now, in order to ``remove'' the indicator from the expectation, we will need to show that, when $\E_T(\delta)$ is false, $\sum_{t\in\tSg} \gradtsq$ cannot be \emph{too large}.
Recall that we have two main tools to upper bound the size of this sum: \cref{lem:trivialGradBound}, which gives a \emph{deterministic} upper bound of $\O(T^3)$, and \cref{lem:logGradBound}, which gives a \emph{high-probability} upper bound of $\tO(T^2)$.
To exploit this ``lighter'' regime of \cref{lem:logGradBound}, it will be useful to introduce the following event:
\begin{align*}
    \E' &= \E_T(\delta)^c \cap \{b_T^2 \leq \constRecImp T^{x + 2} \log^{\gamma_2}(h(T))\},
\end{align*}
where $\delta = \nicefrac{(2+\sigmaOne^2)\log^{y-\gamma_2}(h(T))}{T^{\gamma_1}}$ is the same choice as in \cref{lem:interlacingStepSizeLB}.
By definition, $\E' \subset \E_T(\delta)^c$, so 
\begin{align*}
 \PRO{\E'} \leq \PRO{\E_T(\delta)^c}\leq \delta = \frac{({\color{cblue}2} + \sigmaOne^2) \log^{y-\gamma_2}(h(T))}{T^{\gamma_1}}.    
\end{align*}
Additionally, using Markov's inequality and the assumed upper bound on $\EXP{\sum_{t\in[T]} \gradtsq}$, we may similarly conclude that
\begin{align*}
    \PRO{\E_T(\delta)^c \cap (\E')^c} = \PRO{b_T^2 > \constRecImp T^{x+2}\log^{\gamma_2}(h(T))} \leq \frac{({\color{cblue}2} + \sigmaOne^2)\log^{y-\gamma_2}(h(T))}{T^{2}}.
\end{align*}
Hence, decomposing $\1{\E_T(\delta)} = 1 - \1{\E'} - \1{\E_T(\delta)^c \cap (\E')^c}$, and upper bounding $\sum_{t\in[T]}\gradtsq$ using the \emph{high probability} bound of \cref{lem:logGradBound} under $\E'$, and using the \emph{deterministic} bound of \cref{lem:trivialGradBound} under $\E_T(\delta)^c \cap (\E')^c$, we have that
\begin{align*}
    \EXP{\sum_{t\in\tSg} \gradtsq \1{\E_T(\delta)}}
    &\geq \EXP{\sum_{t\in\tSg} \gradtsq}\\
    &\quad- 2 T (\gradzerosq + \decayBndAGNorm L^2 T \log(\nicefrac{f(T)}{\delta}))\PRO{\E'}\\
                  &\quad- 2 T (\gradzerosq + \stepBndAGNorm^2 L^2 T^2)\PRO{\E_T(\delta)^c \cap (\E')^c}\\
    &\geq \EXP{\sum_{t\in\tSg} \gradtsq}\\
    &\quad- 2 T({\color{cblue}2} + \sigmaOne^2)\log^{y-\gamma_2}(h(T)) \gradzerosq \parens{\frac{1}{T^{\gamma_1}} + \frac{1}{T^2}}\\
    &\quad- 2 T({\color{cblue}2} + \sigmaOne^2)\log^{y-\gamma_2}(h(T)) L^2 \parens{\decayBndAGNorm T^{1-\gamma_1}\log(h(T)) + \stepBndAGNorm^2},
\end{align*}
where in the last inequality, we 
{\color{cblue}use the following facts: $\nicefrac{1}{\delta} = \frac{T^{\gamma_1}}{(2+\sigmaOne^2)\log^{y-\gamma_2}(h(T))} \leq T^2$ (chosen in \cref{lem:interlacingStepSizeLB}) and $\nicefrac{f(T)}{\delta}\leq T^2 f(T)\leq h(T)$ which hold since $\gamma_1\leq 2$, $y-\gamma_2\geq 0$, and by the initial conditions on $h(T)\geq e$.} 
Now, since $\gamma_1 \leq 1$ by assumption (which implies that $2-\gamma_1\geq 1$), we may simplify the above to conclude that
\begin{align*}
    \EXP{\sum_{t\in\tSg} \gradtsq \1{\E_T(\delta)}}
    &\geq \EXP{\sum_{t\in\tSg} \gradtsq}\\
&\quad{\color{cblue}- 4(2+\sigmaOne^2)(\gradzerosq + L^2\eta^2) T^{2-\gamma_1}\log^{y-\gamma_2+1}(h(T)).}
\end{align*}
By our assumption that 
{\color{cblue} $\constRecImp \geq 16(2+\sigmaOne^2)(\gradzerosq + L^2\eta^2)$}, the claimed bound is immediate.
\end{custproof}

\begin{custlemma}[Bounding gradients at the ``bad'' times]\label{lem:interlacingBadTerms}
Recalling the set $\tSg = \Sg \setminus \Sc$ as constructed in \cref{lem:compSetExists}, we have that
\begin{align*}
    \EXP{\sum_{t\not\in\tSg} \gradtsq}
    &\leq 
    \frac{\constRecImp}{2} {\color{cblue}T\log^{2}(h(T)),}
\end{align*}
where $\constRecImp \geq 4(\gradzerosq + \eta^2 L^2)(64(1+\ncomp)\sigmaOne^2  + 1)\1{\sigmaOne>\nicefrac{1}{8}}$ and $h(T) \geq T^2 f(T)$, and $\ncomp = \max\{8\ceil{4\sigmaOne - 1},0\}$.
\end{custlemma}
\begin{custproof}
The main insight of this proof is that, for each time $t$, with high probability, we have that $\gradtsq \leq \O(T\log(h(T)))$, by \cref{lem:logGradBound}. Thus, by using the fact that $\EXP{|\tSgc|} \leq \O(\log(h(T)))$ by \cref{lem:goodSet}, we can hope to obtain the claimed bound. Let us now show how to combine these insights.

First, notice that, by \cref{lem:goodSet}, whenever $\sigmaOne \leq \nicefrac{1}{8}$, then $\Sg = \tSg = [T]$, i.e., $\tSg^c = \emptyset$. Our claim follows trivially in this case. Thus, we will assume for the remainder of the proof that $\sigmaOne > \nicefrac{1}{8}$.

To derive the claimed bound, we will consider the ``nice'' event $\E_T(\delta)$, where $\delta$ is a parameter to be chosen shortly {\color{cblue}(note that $\E_T(\delta)$ need not be the same event as was used in \cref{lem:interlacingStepSizeLB,lem:removingTheIndicator}, as it is simply an event used internally to this proof).}
Recall that we can easily bound $\EXP{|\tSg^c|}$ as:
\begin{align*}
    \EXP{|\tSg^c|}
    = \EXP{|(\Sg\setminus\Sc)^c|}
    = \EXP{|\Sg^c| + |\Sc|}
    &\leq \EXP{(1+\ncomp)|\Sg^c|}\\
    &\leq (1+\ncomp)64\sigmaOne^2\log(f(T)),
\end{align*}
which follows by definition of $\tSg$, together with the construction of $\Sc$ given in \cref{lem:compensationInsight} and the bound on $\EXP{|\Sg^c|}$ from \cref{lem:goodSet}.
Using this fact together with the bounds for $\gradtsq$ from \cref{lem:trivialGradBound,lem:logGradBound}, 
we have that
\begin{align*}
    \EXP{\sum_{t\not\in\tSg} \gradtsq}
    &= \EXP{\sum_{t\not\in\tSg} \gradtsq (\1{\E_T(\delta)} + \1{\E_T(\delta)^c})}\\
    &\leq 2(\gradzerosq + \decayBndAGNorm L^2 T\log(\nicefrac{f(T)}{\delta})) \EXP{|\tSgc|}\\
    &\quad+ 2(\gradzerosq + \stepBndAGNorm^2 L^2T^2) T \PRO{\E_T(\delta)^c}\\
    &\leq \gradzerosq \parens{128(1+\ncomp)\sigmaOne^2 \log(f(T)) + 2 T \delta}\\
    &\quad+ \eta^2 L^2 T \parens{128(1+\ncomp)\sigmaOne^2 \log(\nicefrac{f(T)}{\delta})\log(f(T)) + 2 T^2 \delta}.
\end{align*}
Therefore, choosing $\delta = \nicefrac{1}{T^2}$, and assuming that $T^2 f(T) \leq h(T)$, we conclude that
\begin{align*}
    \EXP{\sum_{t\not\in\tSg} \gradtsq}
    &\leq 2(\gradzerosq + \eta^2 L^2)(64(1+\ncomp) \sigmaOne^2  + 1)\log^2(h(T)) T.
\end{align*}
Since {\color{cblue}$\nicefrac{\constRecImp}{2} \geq 2(\gradzerosq + \eta^2 L^2)(64(1+\ncomp) \sigmaOne^2  + 1)$}, the claimed bound follows from the above.
\end{custproof}

 \section{Obtaining the Convergence Rate for AdaGrad-Norm}
\label{sec:appendix:convergenceAdaGradNorm}

Here, we provide a proof for the main result of this paper, a proof of convergence for the AdaGrad-Norm algorithm.

\begin{custtheorem}\label{thm:affineVariance}
With probability at least $1-\delta$, the AdaGrad-Norm algorithm \eqref{eq:alg} for any choice of parameters $\eta,b_0^2 >0 $ satisfies:
\begin{align}\label{eq:thmSqrtBnd}
    \min_{t\in[T]} \gradtsq
    &\leq \sqrt{1+\sigmaOne^2}\frac{16(\fzero - \fstar + \constInvariance)}{\eta\sqrt{\delta^3 T}}\bigg[b_0 + 2\sigmaZero\nonumber\\
    &\qquad\qquad\qquad\qquad\qquad\qquad\qquad\quad + \sqrt{32(1+8(\ncomp+1)\sigmaOne^2)}(\gradzeronorm + \eta L)\nonumber\\
    &\qquad\qquad\qquad\qquad\qquad\qquad\qquad\quad +16\sqrt{2}\frac{\fzero - \fstar + \constInvariance}{\eta}\bigg] \log^{\nicefrac{13}{4}}(T^2 f(T))\nonumber\\
    &\quad + 
    \frac{\sqrt{2}\parens{128\sigmaOne^2(\ncomp+1)\log(f(T))}^{\frac{3}{2}}{\color{cblue}\gradzerosq}\1{\sigmaOne > \frac{1}{8}}}{(\delta T)^{\nicefrac{3}{2}}},
\end{align}
where $\ncomp = \max\{8\ceil{4\sigmaOne - 1},0\}$ is the constant defined in \cref{lem:compensationInsight},
$\constInvariance$ is the constant defined in \cref{lem:mainSummedBoundAffineInterpreted},
\begin{align*}
    \constInvariance = 2\eta\sigmaZero + \frac{L\eta^2}{2}
    + 128\eta\sigmaOne^2\gradzeronorm\1{\sigmaOne > \nicefrac{1}{8}}
    + L \eta^2 \ncomp(\nicefrac{\ncomp}{8} + 2)(64\sigmaOne^2 + 8192\sigmaOne^4 + 2),
\end{align*}
and
\begin{align*}
    f(T) = {\color{cblue}e} + \frac{\sigmaZero^2 T}{b_0^2} + \frac{(1+\sigmaOne^2 )T}{b_0^2}(\gradzeronorm + {\color{cblue}\eta} L T)^2,
\end{align*}
is the function defined in \eqref{eq:polyFn}.

Furthermore, whenever $\sigmaOne \leq \nicefrac{1}{8}$, then with probability at least $1-\delta$, \eqref{eq:alg} also satisfies:
\begin{align}\label{eq:thmSmallNoiseBound}
    \min_{t\in [T]} \gradtsq
    &\leq \frac{8\sqrt{2}(\fzero - \fstar + \cCommon)}{\delta^2 \eta \sqrt{T}}
    \bigg[\sigmaZero + \sigmaOne \bigg(b_0 + \sigmaZero\nonumber\\
    &\qquad\qquad\qquad\qquad\qquad\qquad\qquad\qquad+ \sqrt{32(1 + 8(1 + \ncomp)\sigmaOne^2)}(\gradzeronorm + \eta L)\nonumber\\
    &\qquad\qquad\qquad\qquad\qquad\qquad\qquad\qquad+ 16\sqrt{2}\frac{\fzero - \fstar + \cCommon}{\eta}\bigg)\bigg]\log^{\nicefrac{9}{4}}(T^2 f(T))\nonumber\\
    &\quad+ \frac{8(\fzero - \fstar + \cCommon \log(f(T)))}{\delta^2 \eta T}\parens{b_0 + 4(2+\sigmaOne^2)\frac{\fzero - \fstar + \cCommon \log(f(T))}{\eta}},
\end{align}
where $\cCommon = 2\eta\sigmaZero + \nicefrac{L\eta^2}{2}$.
\end{custtheorem}

We note that the second bound in \cref{thm:affineVariance}, \eqref{eq:thmSmallNoiseBound}, is particularly interesting in the regime when $\sigmaZero, \sigmaOne = \O(\nicefrac{1}{\sqrt{T}})$. Indeed, in this setting, our bound yields a $\tO(\nicefrac{1}{T})$ convergence rate which one should expect in the noiseless regime.

\begin{custproof}[of \eqref{eq:thmSqrtBnd}]
Now that we know that $\EXP{\sum_{t\in\tSg}\tetat \gradtsq} = \O(\log^2(T))$ by \cref{lem:mainSummedBoundAffineInterpreted} and that $\EXP{\sum_{t\in[T]}\gradtsq} = \tO(T)$ by \cref{lem:recursiveInterlacing}, we have all of the tools we need to obtain our claimed convergence rate. Indeed, we can first use the result of \cref{lem:recursiveInterlacing} to obtain a uniform lower bound on the step size proxies $\tetat$ in expectation. To see this, let us denote 
\begin{align}\label{eq:tetalb}
    \tetalb := \nicefrac{\eta}{\sqrt{b_{T-1}^2 + \sigmaZero^2 + (1+\sigmaOne^2)\sum_{t\in[T]}\gradtsq}},
\end{align}
and observe that $\tetat \geq \tetalb$ for every $t\in[T]$, deterministically. Additionally, by H\"older's inequality, we know that $\EXP{(X Y)^{\nicefrac{2}{3}}} \leq \EXP{X}^{\nicefrac{2}{3}}\EXP{Y^2}^{\nicefrac{1}{3}}$. Thus, taking $X = \tetalb \sum_{t\in\tSg}\gradtsq$ and $Y=\nicefrac{1}{\tetalb},$ we have that
\begin{align}\label{eq:adagradNormPf1}
    \EXP{\sum_{t\in\tSg} \tetat \gradtsq}
    \geq \EXP{\tetalb \sum_{t\in\tSg} \gradtsq}
    \geq \frac{\EXP{\parens{\sum_{t\in\tSg} \gradtsq}^{\frac{2}{3}}}^{\frac{3}{2}}}{\sqrt{\EXP{\parens{\nicefrac{1}{\tetalb}}^2}}}.
\end{align}
To further lower bound \eqref{eq:adagradNormPf1}, we may first upper bound the denominator using our bound from \cref{lem:recursiveInterlacing} together with the definition of $\tetalb$ and \eqref{eq:sgradBound}:
\begin{align*}
    \eta^2\EXP{\parens{\nicefrac{1}{\tetalb}}^2}
    \leq T\sigmaZero^2 + 2(1+\sigmaOne^2)\EXP{\sum_{t\in[T]} \gradtsq}
    \leq T\sigmaZero^2 + 2\constRecImp(1+\sigmaOne^2) T \log^{\frac{5}{2}}(T^2 f(T)).
\end{align*}
Focusing now on lower bounding the numerator of \eqref{eq:adagradNormPf1},
\begin{align*}
    \EXP{\parens{\sum_{t\in\tSg} \gradtsq}^{\frac{2}{3}}}
    = \EXP{|\tSg|^{\frac{2}{3}} \parens{\frac{1}{|\tSg|} \sum_{t\in\tSg} \gradtsq}^{\frac{2}{3}}}
    \geq \EXP{|\tSg|^{\frac{2}{3}} \min_{t\in [T]} \gradtnorm^{\frac{4}{3}}},
\end{align*}
where the lower bound above follows since the average is always larger than the minimum. If it were the case that $\tSg = [T]$, then, at this point, we would essentially be done with our proof. However, since $\tSg$ is a random set, we must take some additional care. Because $|\tSgc|$ is $\O(\log(T))$ in expectation by \cref{lem:goodSet}, this is only a minor technicality. Indeed,
\begin{align*}
    \EXP{|\tSg|^{\frac{2}{3}} \min_{t\in [T]} \gradtnorm^{\frac{4}{3}}}
    \geq\parens{\frac{T}{2}}^{\frac{2}{3}}\EXP{\min_{t\in [T]} \gradtnorm^{\frac{4}{3}}\1{|\tSg|\geq \nicefrac{T}{2}}}.
\end{align*}
Therefore, collecting the results we have derived so far into a lower bound on the right-hand side of \eqref{eq:adagradNormPf1}, and applying the result of \cref{lem:mainSummedBoundAffineInterpreted} to upper bound the left-hand side of \eqref{eq:adagradNormPf1}, we have obtained the following upper bound:
\begin{align}\label{eq:adagradNormPf2}
    \EXP{\min_{t\in[T]} \gradtnorm^{\frac{4}{3}} \1{|\tSg|\geq\nicefrac{T}{2}}}
    \leq \parens{\frac{C_T}{\sqrt{T}}}^{\frac{2}{3}},
\end{align}
where 
\begin{align*}
    C_T := \frac{{\color{cblue}8}(\fzero - \fstar + \constInvariance\log^2(T^2 f(T)))}{\eta} \sqrt{\sigmaZero^2 + 2 \constRecImp(1+\sigmaOne^2) \log^{\nicefrac{5}{2}}(T^2 f(T))}.
\end{align*}
To conclude, we will translate the bound in \eqref{eq:adagradNormPf2} into one on $\EXP{\min_{t\in[T]} \gradtnorm^{\nicefrac{4}{3}}}$. Begin by writing
\begin{align*}
    \EXP{\min_{t\in[T]} \gradtnorm^{\frac{4}{3}}}
    &=\EXP{\min_{t\in[T]} \gradtnorm^{\frac{4}{3}}\1{|\tSg| \geq \nicefrac{T}{2}}}
    + \EXP{\min_{t\in[T]} \gradtnorm^{\frac{4}{3}}\1{|\tSgc| \geq \nicefrac{T}{2}}}\\
    &\leq \parens{\frac{C_T}{\sqrt{T}}}^{\frac{2}{3}}
    + \gradzeronorm^{\frac{4}{3}}\PRO{|\tSgc| \geq \nicefrac{T}{2}}.
\end{align*}
where the inequality follows since $\min_{t\in[T]}\gradtnorm^{\nicefrac{4}{3}} \leq \gradzeronorm^{\nicefrac{4}{3}}$.
The above failure probability can be easily upper bounded via Markov's inequality:
\begin{align*}
    \PRO{|\tSgc| \geq \nicefrac{T}{2}}
    &\leq \frac{2(\ncomp+1) \EXP{|\Sg^c|}}{T}
    \leq \frac{128\sigmaOne^2(\ncomp+1)\log(f(T))\1{\sigmaOne > \nicefrac{1}{8}}}{T},
\end{align*}
where we used the fact that, by \cref{lem:compSetExists}, $|\tSgc| = |\Sg^c \cup \Sc| = (\ncomp + 1)|\Sg^c|$, along with \cref{lem:goodSet}.
The above bound combined with \eqref{eq:adagradNormPf2} thus gives
\begin{align*}
    \EXP{\min_{t\in[T]} \gradtnorm^{\frac{4}{3}}} \leq \parens{\frac{C_T}{\sqrt{T}}}^{\frac{2}{3}} + \frac{128\sigmaOne^2(\ncomp+1)\log(f(T)) {\color{cblue} \gradzeronorm^{\nicefrac{4}{3}}}\1{\sigmaOne > \frac{1}{8}}}{T}
    \leq \parens{\frac{\widetilde{C}_T}{\sqrt{T}}}^{\frac{2}{3}},
\end{align*}
where 
\begin{align*}
    \widetilde{C}_{T} 
    &:= \sqrt{2}\parens{C_T + \frac{\parens{128\sigmaOne^2(\ncomp+1)\log(f(T))}^{\frac{3}{2}}{\color{cblue}\gradzerosq}\1{\sigmaOne > \frac{1}{8}}}{T}}\\
    &\geq \parens{C_T^{\frac{2}{3}} + \frac{128\sigmaOne^2(\ncomp+1)\log(f(T)) {\color{cblue} \gradzeronorm^{\nicefrac{4}{3}}}\1{\sigmaOne > \frac{1}{8}}}{T^{\nicefrac{2}{3}}}}^{\frac{3}{2}}.
\end{align*}
Hence, by a final application of Markov's inequality, we obtain, for any $\delta\in(0,1)$,
\begin{align*}
    \PRO{\min_{t\in[T]} \gradtsq > \frac{\widetilde{C}_T}{\sqrt{\delta^3 T}}}
    = \PRO{\min_{t\in[T]} \gradtnorm^{\frac{4}{3}} > \frac{1}{\delta}\parens{\frac{\widetilde{C}_T}{\sqrt{T}}}^{\frac{2}{3}}}
    \leq \delta.
\end{align*}
as claimed.
\end{custproof}

\begin{custproof}[of \eqref{eq:thmSmallNoiseBound}]
{\color{cblue}
We will proceed in a similar manner as in the proof of \eqref{eq:thmSqrtBnd}, borrowing notation from that proof, and using a slightly different application of H\"older's inequality, which will allow us to prove a $\tO(\nicefrac{1}{T})$ convergence rate in the ``small-noise'' regime. We begin by noting that, whenever $\sigmaOne \leq \nicefrac{1}{8}$, $\Sg = \tSg = [T]$ by \cref{lem:goodSet}. Thus, for the purpose of this proof, we will replace $\tSg$ with $[T]$.

Using the fact $\sum_{t\in \tSg}\gradtsq = \sum_{t\in [T]}\gradtsq$ in our setting ($\sigmaOne \leq \nicefrac{1}{8}$), we apply H\"older's inequality $\EXP{\sqrt{X Y}} \leq \sqrt{\EXP{X} \EXP{Y}}$, where we choose $X$ and $Y$ as $X = \nicefrac{1}{\tetalb}$ and $Y = \tetalb \sum_{t\in[T]} \gradtsq$, to establish that
\begin{align*}
    \EXP{\sum_{t\in \tSg} \tetat\gradtsq}
    \geq \EXP{\tetalb \sum_{t\in[T]} \gradtsq}
    \geq \frac{\EXP{\sqrt{\sum_{t\in [T]} \gradtsq}}^2}{\EXP{\nicefrac{1}{\tetalb}}}\\
\end{align*}
Now, plugging in the definition of $\tetalb$ from \eqref{eq:tetalb} and upper-bounding $\sgradtsq \leq 2(\norm{\sgradt - \gradt}^2 + \gradtsq)$, we can conclude that
\begin{align*}
    \EXP{\nicefrac{\eta}{\tetalb}}
    &\leq \EXP{\sqrt{b_0^2 + \sigmaZero^2 + 2\sum_{t\in[T-1]}\norm{\sgradt - \gradt}^2}} + \EXP{\sqrt{(2+\sigmaZero^2)\sum_{t\in[T]}\gradtsq}}\\
    &\leq \sqrt{b_0^2 + {\color{cmaroon}2}T\sigmaZero^2 + 2\sigmaOne^2 \constRecImp T \log^{\nicefrac{\color{cmaroon}5}{2}}(T^2 f(T))} + \EXP{\sqrt{(2+\sigmaZero^2)\sum_{t\in[T]}\gradtsq}},
\end{align*}
where the second inequality follows by Jensen's inequality together with \cref{assump:affineVariance} and the bound on $\EXP{\sum_{t\in[T]} \gradtsq}$ from \cref{lem:recursiveInterlacing}. Therefore, using the result of \cref{lem:mainSummedBoundAffineInterpreted} to upper-bound $\EXP{\sum_{t\in[T]}\tetat \gradtsq}$ together with the lower bound on the same quantity that we have just derived above, and writing $a = \sqrt{b_0^2 + 2T\sigmaZero^2 + 2\sigmaOne^2 \constInvariance T \log^{\nicefrac{5}{2}}(T^2 h(T))}$ and $b = \nicefrac{4(\fzero - \fstar + \cCommon \log(f(T)))}{\eta}$, we conclude that
\begin{align*}
    \EXP{\sqrt{\sum_{t\in [T]} \gradtsq}}^2 \leq b\parens{a + \sqrt{2+\sigmaOne^2}\EXP{\sqrt{\sum_{t\in [T]} \gradtsq}}}.
\end{align*}
Solving this quadratic inequality, we conclude that
\begin{align*}
    \EXP{\sqrt{\sum_{t\in [T]} \gradtsq}} 
    \leq \frac{\sqrt{2+\sigmaOne^2} b}{2} + \sqrt{\frac{(2 + \sigmaOne^2) b^2}{4} + a b}
    \leq \sqrt{2+\sigmaOne^2} b + \sqrt{a b}.
\end{align*}
Now, using the fact that 
\begin{align*}
    \EXP{\sqrt{\sum_{t\in [T]} \gradtsq}} = \sqrt{T}\EXP{\sqrt{\frac{1}{T}\sum_{t\in [T]} \gradtsq}} \geq \sqrt{T}\EXP{ \min_{t\in[T]}\gradtnorm},
\end{align*}
we conclude that
\begin{align*}
    \EXP{\min_{t\in [T]}\gradtnorm} \leq \frac{\sqrt{2+\sigmaOne^2} b + \sqrt{a b}}{\sqrt{T}}.
\end{align*}
In particular, this implies by Markov's inequality that, with probability at least $1-\delta$,
\begin{align*}
    \min_{t\in [T]} \gradtsq
    &\leq \frac{2(2 + \sigmaOne^2)b^2 + 2 a b}{\delta^2 T}\\
    &\leq \frac{8(\fzero - \fstar + \cCommon \log(f(T)))}{\delta^2 \eta T}\parens{b_0 + 4(2+\sigmaOne^2)\frac{\fzero - \fstar + \cCommon \log(f(T))}{\eta}}\\
    &\quad +\frac{8\sqrt{2}(\fzero - \fstar + \cCommon \log(f(T)))}{\delta^2 \eta \sqrt{T}} \sqrt{\sigmaZero^2 + \constRecImp \sigmaOne^2 \log^{\nicefrac{5}{2}}(T^2 f(T))}.
\end{align*}
{\color{cmaroon}T}his shows that, in the setting when $\sigmaZero^2, \sigmaOne^2 = \O(\nicefrac{1}{\sqrt{T}})$, then we recover a $\tO(\nicefrac{1}{T})$ convergence rate, as in the noiseless setting.
}
\end{custproof}
 
\end{document}